\documentclass[letterpaper, 10 pt, conference]{IEEEtran}
\IEEEoverridecommandlockouts
\usepackage{cite}
\usepackage{amsmath,amssymb,amsfonts}
\usepackage{algorithmic}
\usepackage{graphicx}
\usepackage{textcomp}
\usepackage{xcolor}
\usepackage{flushend}
\def\BibTeX{{\rm B\kern-.05em{\sc i\kern-.025em b}\kern-.08em
    T\kern-.1667em\lower.7ex\hbox{E}\kern-.125emX}}

\usepackage{xcolor}
\newcommand{\cmt}[1]{}

\long\def\ignorethis#1{}

\newcommand{\vc}[1]{\ensuremath{\mathbf{#1}}}

\newcommand{\mat}[1]{\ensuremath{\mathbf{#1}}}

\newcommand{\pctab}{\hspace{0.2in}}

\begin{document}

\title{\LARGE \bf Learning a Single Policy for Diverse Behaviors on a Quadrupedal Robot using Scalable Motion Imitation
}

\author{Arnaud Klipfel$^{1}$ Nitish Sontakke$^{1}$ Ren Liu$^{2}$ Sehoon Ha$^{1}$
\thanks{$^{1}$School of Interactive Computing, Georgia Institute of Technology, Atlanta, GA, 30308, USA. {\tt \small aklipfel3@gatech.edu, nitishsontakke@gatech.edu, sehoonha@gatech.edu}}%
\thanks{$^{2}$ Meta Platforms, Inc., USA, 
        {\tt\small renl@meta.com}. Work done while at Georgia Tech.}%
}

\maketitle

\begin{abstract}
Learning various motor skills for quadrupedal robots is a challenging problem that requires careful design of task-specific mathematical models or reward descriptions. In this work, we propose to learn a single capable policy using deep reinforcement learning by imitating a large number of reference motions, including walking, turning, pacing, jumping, sitting, and lying. On top of the existing motion imitation framework, we first carefully design the observation space, the action space, and the reward function to improve the scalability of the learning as well as the robustness of the final policy. In addition, we adopt a novel adaptive motion sampling (AMS) method, which maintains a balance between successful and unsuccessful behaviors. This technique allows the learning algorithm to focus on challenging motor skills and avoid catastrophic forgetting. We demonstrate that the learned policy can exhibit diverse behaviors in simulation by successfully tracking both the training dataset and out-of-distribution trajectories. We also validate the importance of the proposed learning formulation and the adaptive motion sampling scheme by conducting experiments. 
\end{abstract}


\section{Introduction}

Quadrupedal robots can achieve various autonomous missions by overcoming rough terrains that wheeled robots cannot traverse, but the control is not straightforward due to its high-dimensional state space and under-actuated dynamics. Roboticists have studied various approaches for legged robot control, ranging from model-based control~\cite{raibert1986legged,park2017high,bledt2018cheetah,kim2019highly} to learning-based approaches~\cite{hwangbo2019learning,lee2020learning,kumar2021rma,miki2022learning}, which have demonstrated impressive agility and robustness on various quadrupedal robots. However, most of the prior works have focused on the given specific task, such as robust walking, running, or jumping, because they are governed by very different dynamics. These task-specific controllers often require manual engineering based on the expert's prior knowledge, which can be either developing mathematical models for model-based controllers or shaping reward functions for learning-based algorithms. It requires even more effort if the developer wants to improve the naturalness of the behavior.

One interesting approach is to develop a motion imitation controller that can track the given reference motion, which defines the task implicitly. For instance, walking and jumping are two very different tasks, but motion imitation treats them as the same task of tracking the corresponding motion. If the reference is captured by a human or an animal, motion imitation can also allow us to develop natural behaviors from the original actor. One notable early work is DeepMimic proposed by Peng et al.~\cite{deepmimic}, which shows an impressive motion-tracking performance on a simulated humanoid character. Many researchers have extended this work to imitate a wide range of motions on a simulated character by investigating novel policy architectures~\cite{won2020} or introducing adversarial learning~\cite{peng2021amp,peng2022ase}. This motion imitation approach has been investigated in the context of robotics as well to develop natural motions~\cite{peng2020learning}, but it is limited to tracking a single reference motion.

\begin{figure}
    \vspace{5mm}

\centerline{\includegraphics[width=0.45\textwidth]{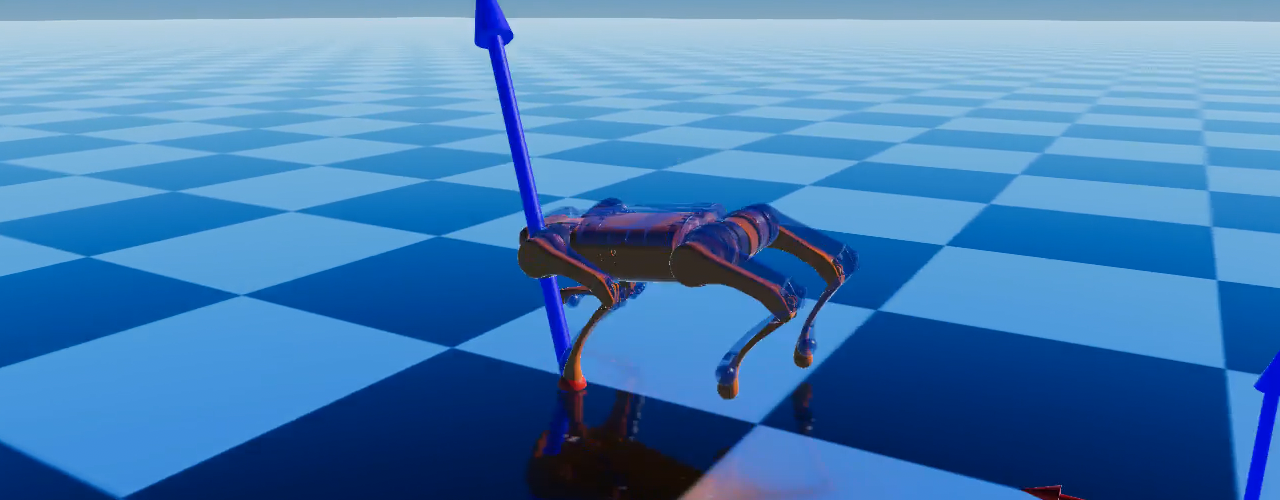}}
\centerline{\includegraphics[width=0.45\textwidth]{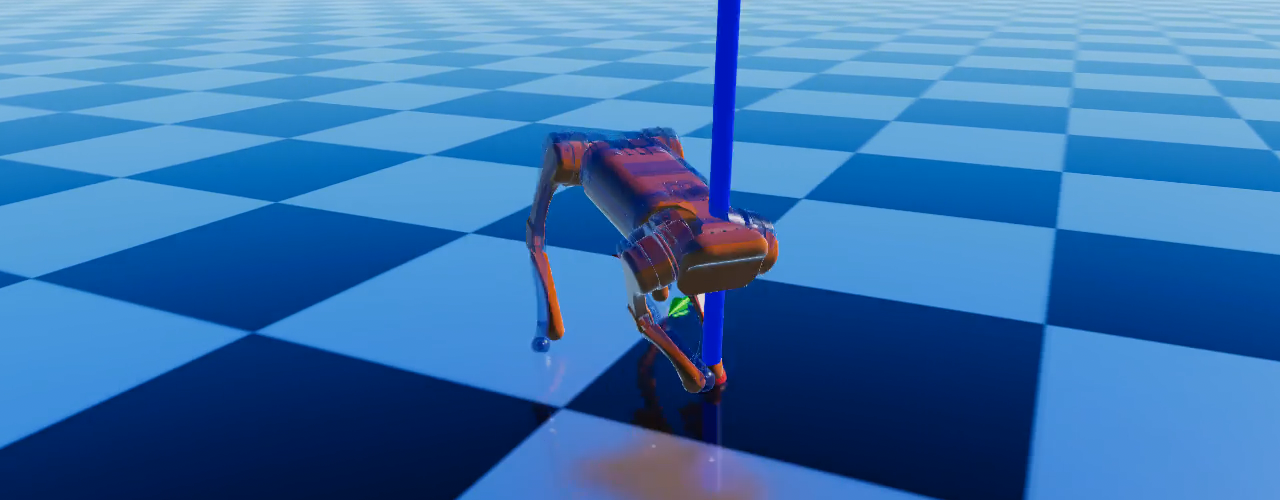}}
\centerline{\includegraphics[width=0.45\textwidth]{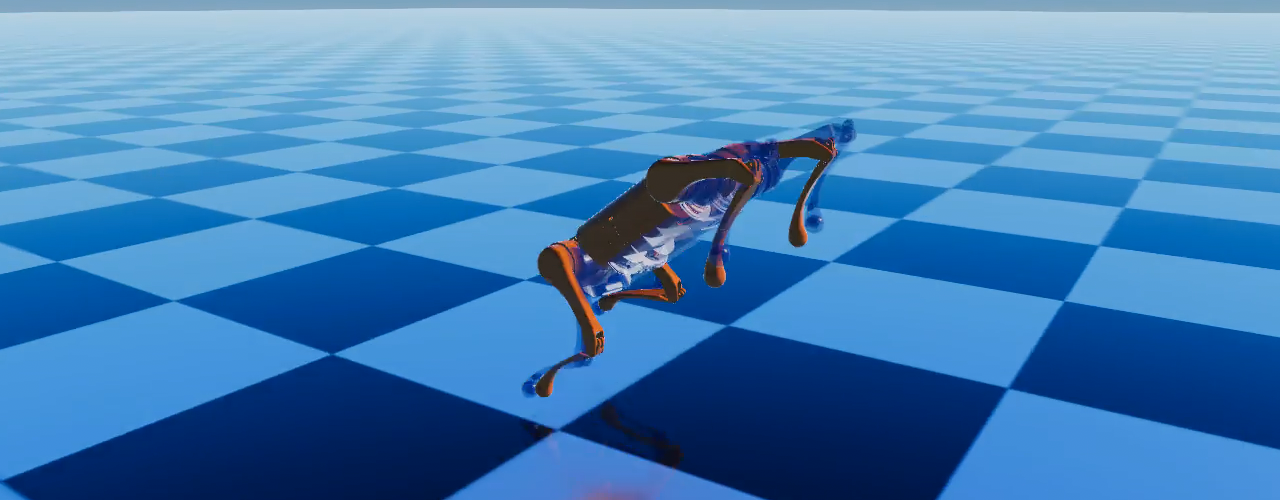}}
\centerline{\includegraphics[width=0.45\textwidth]{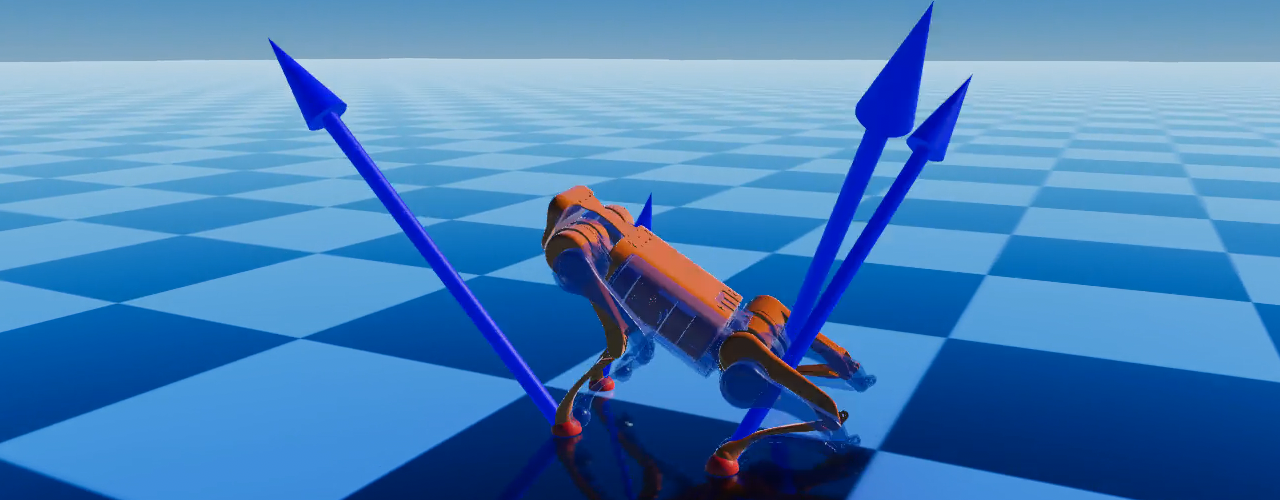}}

\caption{Our scalable motion imitation framework can learn a single policy to execute many motor skills, from running (\textbf{1st}), turning (\textbf{2nd}), jumping (\textbf{3rd}), and sitting (\textbf{4th}).}
\label{fig-teaser}
\end{figure}

This paper investigates a scalable motion imitation framework for a quadrupedal robot to track various motor skills using a single control policy. For the preprocessing, we carefully prepare all the reference motions by retargeting the existing dog's motion database, which results in $701$ motion clips with $15$ different motion types. Then we extend the existing motion imitation framework~\cite{deepmimic} to improve its scalability and robustness. We design a new problem formulation, including a new observation space that includes future and past references and a new reward function that does not incentivize low-level kinematic tracking. In addition, we propose a novel adaptive motion sampling (AMS) scheme to learn all the trajectories without ignoring some outlier motions, such as jumping, and also to avoid catastrophic forgetting about the previously learned motor skills. 

We demonstrate that our framework can learn a single versatile motion imitation policy that can track a large variety of reference motions. Our policy can even track new out-of-distribution trajectories, such as a star-shaped path or a motion with multiple jumps. By conducting an ablation study, we show that adaptive trajectory sampling is necessary to learn all the motor skills in the database. We also demonstrate the robustness of the framework, which is achieved by our novel learning formulation. Our key contributions are summarized as follows:
\begin{itemize}
\item We propose novel techniques for greatly improving the robustness and scalability of motion imitation.
\item We showcase that our framework can learn a single policy to track various challenging trajectories.
\item We validate the proposed components by conducting ablation studies.
\end{itemize}



\section{Related works}

\subsection{Quadrupedal Locomotion}
The control of quadrupedal robots has been thoroughly studied by many robotics researchers. One common approach is to develop a model-based controller that captures the important characteristics of the robot's dynamics using a mathematical model and generates optimal control trajectories~\cite{raibert1986legged,park2017high,bledt2018cheetah,kim2019highly}. While demonstrating impressive robustness and agility on hardware, a model-based approach often requires manual engineering to develop the proper dynamics model for the given task. In recent years, researchers have showcased that it is possible to learn robust locomotion policies using deep reinforcement learning (deep RL)~\cite{hwangbo2019learning,lee2020learning,kumar2021rma,miki2022learning}. However, it is also a well-known challenge that deep RL often requires an extensive amount of reward shaping to obtain the best quality policy that can be effectively transferred to the real world. Therefore, the developed reward functions sometimes have many different terms, up to nine or ten, to guarantee symmetric, energy-efficient, cyclic, and effective gaits~\cite{rudin2022learning,margolis2022rapid}. Therefore, developing a high-quality motion controller for novel tasks still remains a challenging problem for both model-based and learning-based approaches and requires a lot of human effort.

\subsection{Motion Imitation}
Motion imitation is a problem formulation that aims to track the given reference motion. Because the task is implicitly encoded in the reference, this framework allows us to use a unified problem formulation for various tasks, unlike standard task-based problem formulations. The early work of Peng et al.~\cite{deepmimic} demonstrates that it is possible to train a virtual human character to track a single motion in a physics-based simulation. This research is followed by many other works in computer animation~\cite{won2020,won2021control,peng2021amp,peng2022ase} to track a wide range of motions. The robotic community also adopts the same motion imitation framework to develop quadrupedal robot controllers to achieve natural animal-looking motions~\cite{peng2020learning}. Kim et al.~\cite{kim2022human} demonstrate a human motion interface to control a quadrupedal robot by combining motion imitation and motion retargeting. Escontrela et al.~\cite{escontrela2022adversarial} show that adversarial reward formulation of motion imitation can be a good substitute option for complex reward functions. Our work is also closely related to these state-of-the-art contributions in both the computer animation and robotics communities. We extend the motion imitation framework to support a large dataset for quadrupedal locomotion by proposing a novel adaptive sampling and policy design.

\section{Scalable Motion Imitation}
In this section, we will describe the proposed scalable motion imitation framework to track more than $700$ motion clips as well as out-of-distribution trajectories with a single policy. We first explain our data generation procedure in Section~\ref{sec:data_generation}. Then we present our novel problem formulation in Section~\ref{sec:problem}, which is designed to improve the robustness of the existing motion imitation framework. Finally, we describe our novel adaptive trajectory sampling method in Section~\ref{sec:adaptive_sampling}, which is necessary to learn a large number of motion trajectories.

\subsection{Data Generation} \label{sec:data_generation}

\begin{figure}
    \vspace{5mm}

\centerline{\includegraphics[width=0.46\textwidth]{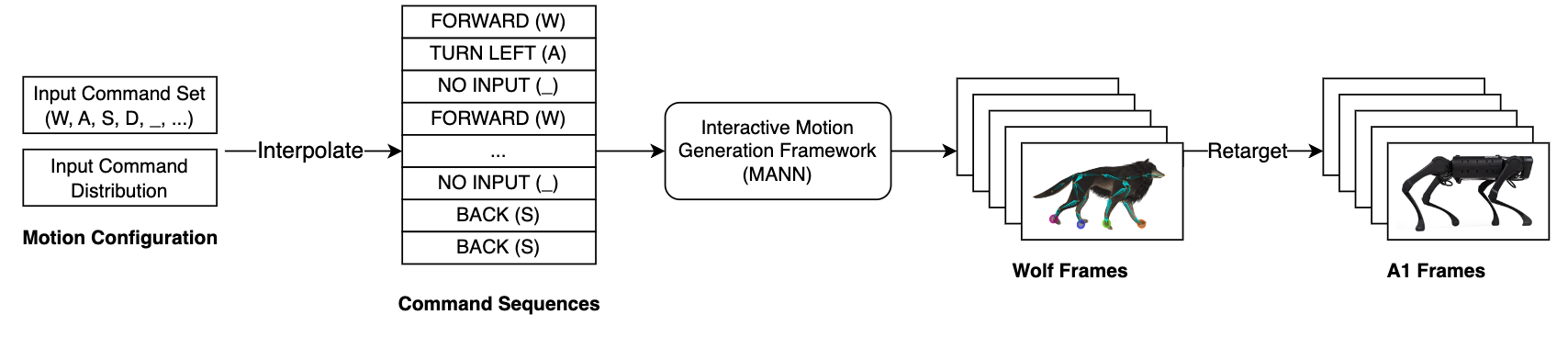}}
\caption{Motion dataset generation pipeline.}
\label{fig-dataset-gen-pipeline}
\end{figure}

The motion dataset contains $701$ motion clips with $15$ different motion types. We generate the dataset using our motion generation pipeline (See Fig.~\ref{fig-dataset-gen-pipeline}). Every motion clip in the dataset lasts for $10$ seconds and is of $60$~Hz frame rate. For the random keyboard input commands and their distribution, the pipeline interpolates them into a sequence of 600 to simulate user interaction. Then we infer kinematic data, such as joint angles, with the interactive motion generation framework~\cite{zhang2018mode}, which is trained with a wolf character skeleton. 

\begin{figure}
    \vspace{5mm}

\centerline{\includegraphics[width=0.3\textwidth]{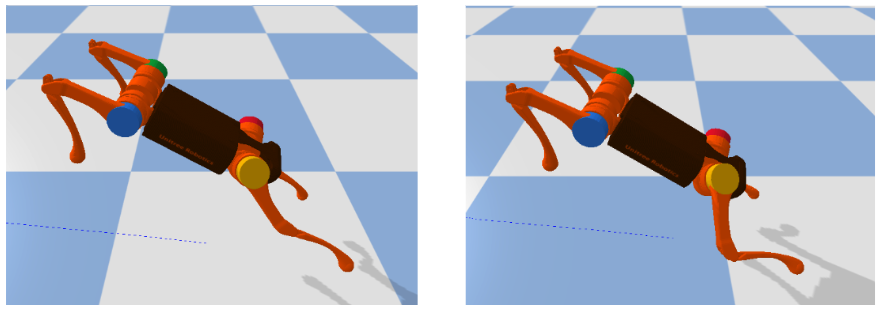}}
\caption{Jumping with different scaling factors.(\textbf{Left}:  1.0, \textbf{Right}: 0.825 (ours)).}
\label{fig-dataset-jump-comp}
\end{figure}

\begin{figure}
\centerline{\includegraphics[width=0.3\textwidth]{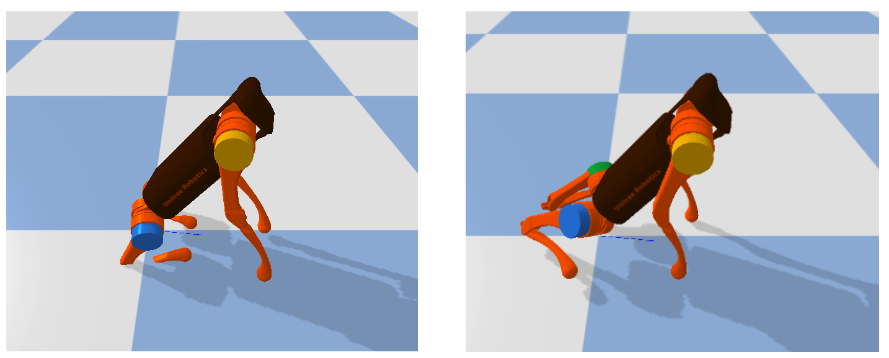}}
\caption{Different retargeted sit motions. \textbf{Left:} IK with hips and feet restriction. \textbf{Right (ours)}: Adjusted IK considering knee positions for sit motions.}
\label{fig-dataset-sit-comp}
\end{figure}

Therefore, we need to retarget the motion of a wolf into our A1 quadrupedal robot~\cite{a1}. We apply the algorithm implemented by Peng et al. \cite{peng2020learning}, which pairs corresponding key points from the source subject's body to the target robot's body, including the positions of the feet and hips, and then performs inverse-kinematics (IK) \cite{gleicher1998retargetting} to fulfill the morphological gap. The pipeline has many robot model-specific hyperparameters that affect the results. For instance, Fig. \ref{fig-dataset-jump-comp} shows that an improper coordinates scaling may cause physically unreasonable front knee joint angles when jumping. We also need to remove all the ground penetration: therefore, our retargeting algorithm is aware of the motion type to handle such special cases (See Fig. \ref{fig-dataset-sit-comp}). We further remove some artifacts, such as foot skating or jitterness, by applying inverse kinematics and smoothing. The content of the generated database is summarized in Table~\ref{tab1-dataset}.

\begin{table}
\caption{Generated Motion Clip Dataset Specification}
\begin{center}
\begin{tabular}{|c|c|c|c|c|c|c|}
\hline
\textbf{Motion Type} & Stand & Step & Pace & Trot & Gallop & Jump \\
\hline
\textbf{Number of Clips} & 1 & 1 & 200 & 50 & 1 & 200 \\
\hline
\hline
\textbf{Motion Type} & \multicolumn{2}{|c|}{Turn Left} & \multicolumn{2}{|c|}{Turn Right} & Sit & Lie\\
\hline
\textbf{Number of Clips} & \multicolumn{2}{|c|}{51} & \multicolumn{2}{|c|}{51} & 1 & 1 \\
\hline
\hline
\textbf{Motion Type} & \multicolumn{3}{|c|}{Turn Left In-Place} & \multicolumn{3}{|c|}{Turn Right In-Place} \\
\hline
\textbf{Number of Clips} & \multicolumn{3}{|c|}{25} & \multicolumn{3}{|c|}{25} \\
\hline
\hline
\textbf{Motion Type} & \multicolumn{2}{|c|}{Triangle Trace} & \multicolumn{2}{|c|}{Star Trace} & \multicolumn{2}{|c|}{Random Mixed} \\
\hline
\textbf{Number of Clips} & \multicolumn{2}{|c|}{1} & \multicolumn{2}{|c|}{1} & \multicolumn{2}{|c|}{92}  \\
\hline
\end{tabular}
\label{tab1-dataset}
\end{center}
\end{table}

\subsection{Problem Formulation for Scalability and Robustness} \label{sec:problem}
Imitating the given reference motion is a popular approach to developing a versatile physics-based controller. While researchers traditionally have approached this problem using model-based control~\cite{li2021fastmimic}, the recent advances in deep reinforcement learning offer an automated approach to learning a tracking policy for a variety of motions. We mostly follow the formulation of Peng et al.~\cite{deepmimic}, while making a few adjustments in the state, action, and reward function designs. We define our problem as a Markov Decision Process with reward function at time $t$, $r_t$, action $\vc{a}_t$, observation $\vc{o}_t$ and state $\vc{s}_t$. 

\subsubsection{Observation Space}
In our formulation, the observation space consists of three components: the robot proprioceptive data, some privileged data, and the reference motion data to track. For each control time step $t\in\mathbb{R}$ (every $0.02s$), the robot proprioception is composed of the joint positions in radians $\vc{q}_t\in\mathbb{R}^{12}$, the joint velocities $\dot{\vc{q}}_t\in\mathbb{R}^{12}$ in rad/s which are given by the encoders, the angular velocity $\omega_t\in\mathbb{R}^{3}$ in rad/s which is given by the robot on-board gyroscope. The policy has also access to some privileged information that is usually not estimated on a robot, or which requires some additional estimation than just pure proprioceptive readings \cite{Ji_2022}. This privileged information is composed of the Center-Of-Mass (CoM) $\vc{x}_t\in\mathbb{R}^{3}$ of the robot with respect to (w.r.t) an origin frame (the same as the reference data), the robot base orientation w.r.t the same origin frame given as the full rotation matrix $\vc{R}_t\in SO(3)$, and the body linear velocity at the CoM $\vc{v}_t\in\mathbb{R}^{3}$, expressed in the inertial frame of the robot. The CoM position could be estimated using a joint or visual odometry, as well as the rotation matrix, and the velocity could also be estimated using the accelerometer data \cite{Ji_2022}. The robot data is written as: $\vc{o}^{robot}_t=(\vc{x}_t^T,(\vc{R}_t)_{i,j}^{i,j\in[1,3]}, \vc{q}_t^T, \vc{v}_t^T,\omega_t^T,\dot{\vc{q}}_t^T)\in \mathbb{R}^{42}$.  $(\vc{A})_{i,j}$ refers to the coefficient $(i,j)$ of the matrix $\vc{A}$.
\\\indent In contrast to other works~\cite{deepmimic, won2020}, the observation space does not include the state of every joint and link of the robot (i.e. twist information, orientation, and relative body position w.r.t. the root joint) and contains only the base full state and the joint angles. The policy has not access to a key frame identifier or marker such as a normalized phase variable \cite{deepmimic} that is used to make the motion learning faster. 
\\\indent The reference motion vector $\bar{\vc{m}}_t \in \mathbb{R}^{24\times8=192}$ is comprised of the target joint positions $\bar{\vc{q}}_t\in\mathbb{R}^{12}$, the rotation matrix $\bar{\vc{R}}_t\in SO(3)$, and the CoM position w.r.t an origin frame $\mathbf{\bar{\vc{x}}}_t\in\mathbb{R}^{3}$. Then our observation $\vc{o}_t \in \mathbb{R}^{234}$ is defined as the concatenation of the robot data and the reference motion data for a short time window, $\vc{o}_t = ({\vc{o}^{robot}_t}^T, \bar{\vc{m}}_{t-1.0}^T, \bar{\vc{m}}_{t-0.5}^T ,  \bar{\vc{m}}_{t-0.2}^T, \bar{\vc{m}}_{t-0.02}^T, \allowbreak \bar{\vc{m}}_{t+0.02}^T, \bar{\vc{m}}_{t+0.2}^T, \bar{\vc{m}}_{t+0.5}^T, \bar{\vc{m}}_{t+1.0}^T\allowbreak)$. Note that we include both past and future reference motions for learning efficiency. We also exclude the current reference frame $\bar{\vc{m}}_t$ to avoid the copy-and-paste behavior of the current frame and promote broader exploration, i.e. adaptation of the low-level joint positions of the reference to the robot and environment dynamics. 

\subsubsection{Action Space}
The action $\vc{a}_t \in \mathbb{R}^{12}$ is defined as the delta to a nominal (i.e independent of the reference and fixed at all time) joint configuration of the robot, which becomes the target position for the proportional-derivative controller at each joint. The generated actions are further smoothed by applying a moving average with a window size of two. The nominal joint configuration is: $\vc{a}_m=(                -0.01, 0.75, -1.5,
                0.01, 0.75, -1.5,
                -0.01, 0.75, -1.5,\allowbreak
                0.01, 0.75, -1.5)$, which corresponds to a standing configuration. Joint positions are bounded. $\theta_{hip}\in[-0.5,0.5] rad$, $\theta_{thigh}\in[-0.1,1.5] rad$, and $\theta_{calf}\in[-2.1,-0.5] rad$. 
\subsubsection{Reward Function} \label{sec:reward}
We design our reward function as follows:
\begin{align} \label{eq:reward}
    r_t &=  w_1 \textit{exp}({-k_1 || \bar{\vc{x}}_t - \vc{x}_t||^2}) + w_2 \textit{exp}({-k_2 || \bar{\mat{R}}_t - \mat{R}_t||^2}) \nonumber \\
    &+ w_3 \textit{exp}({-k_3 || \bar{\vc{e}}_t - \vc{e}_t||^2}),
\end{align}
where $\vc{x}$, $\mat{R}$, and $\vc{e}$ are the root position, the base orientation represented as a rotation matrix, and the end-effector positions expressed in the origin frame. The other terms $\bar{\vc{x}}$, $\bar{\vc{R}}$, and $\bar{\vc{e}}$ are the corresponding desired values from the reference motions. Therefore, each term encourages to track the given reference motion. $w_1$, $w_2$, and $w_3$ are the weight vectors to adjust the importance of each term and $k_1$, $k_2$, and $k_3$ are additional decaying parameters to tune the sensitivity of the reward term. We set the parameters as $w_1=0.7$, $w_2 = 0.5$, $w_3 = 0.15$, $k_1 = 12.5$, $k_2 = 20.0$, and $k_3 = 40.0$ for all the experiments.

As discussed in \ref{sec:reward}, we do not have a low-level tracking reward term as \cite{won2020, deepmimic, peng2020learning}, in order to prevent overfitting on the kinematics, and as such we provide more joint position information in the observation space to passively enforce the reference joint positions. 

\begin{figure*}
    \vspace{5mm}

    \centering
    \setlength{\tabcolsep}{1pt}
    \renewcommand{\arraystretch}{0.7}
    \begin{tabular}{c c c c c}
    \includegraphics[width=0.195\textwidth]{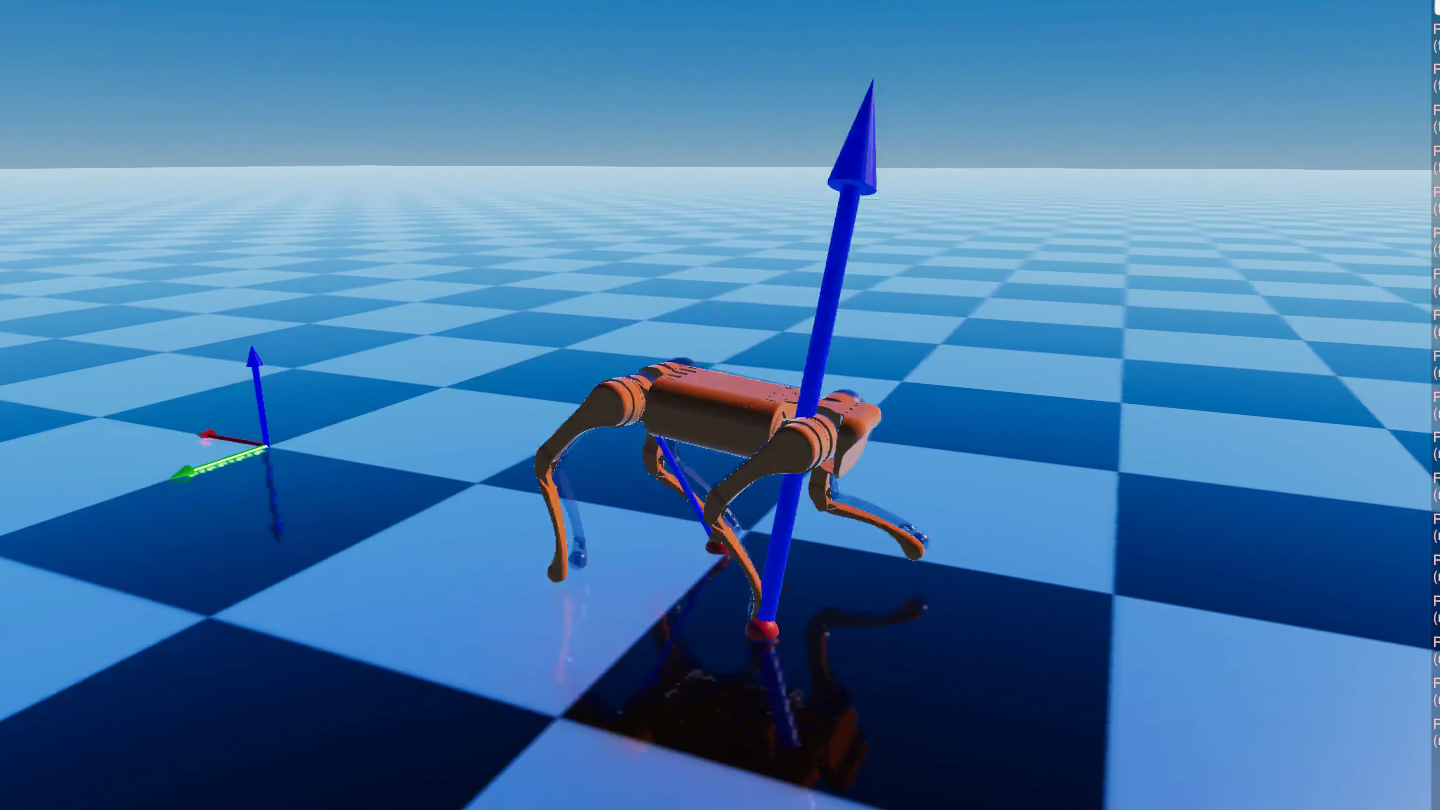} &
    \includegraphics[width=0.195\textwidth]{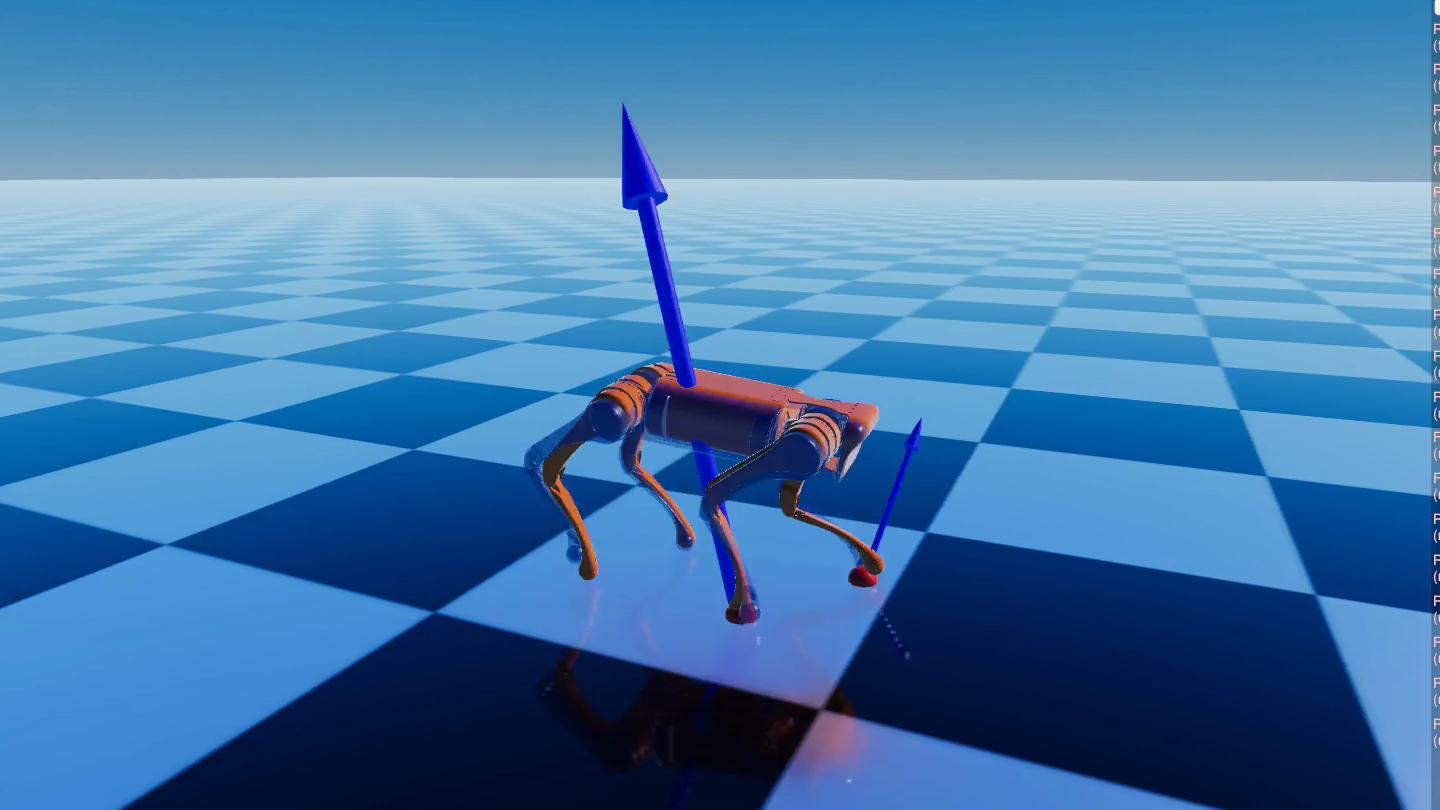} &
    \includegraphics[width=0.195\textwidth]{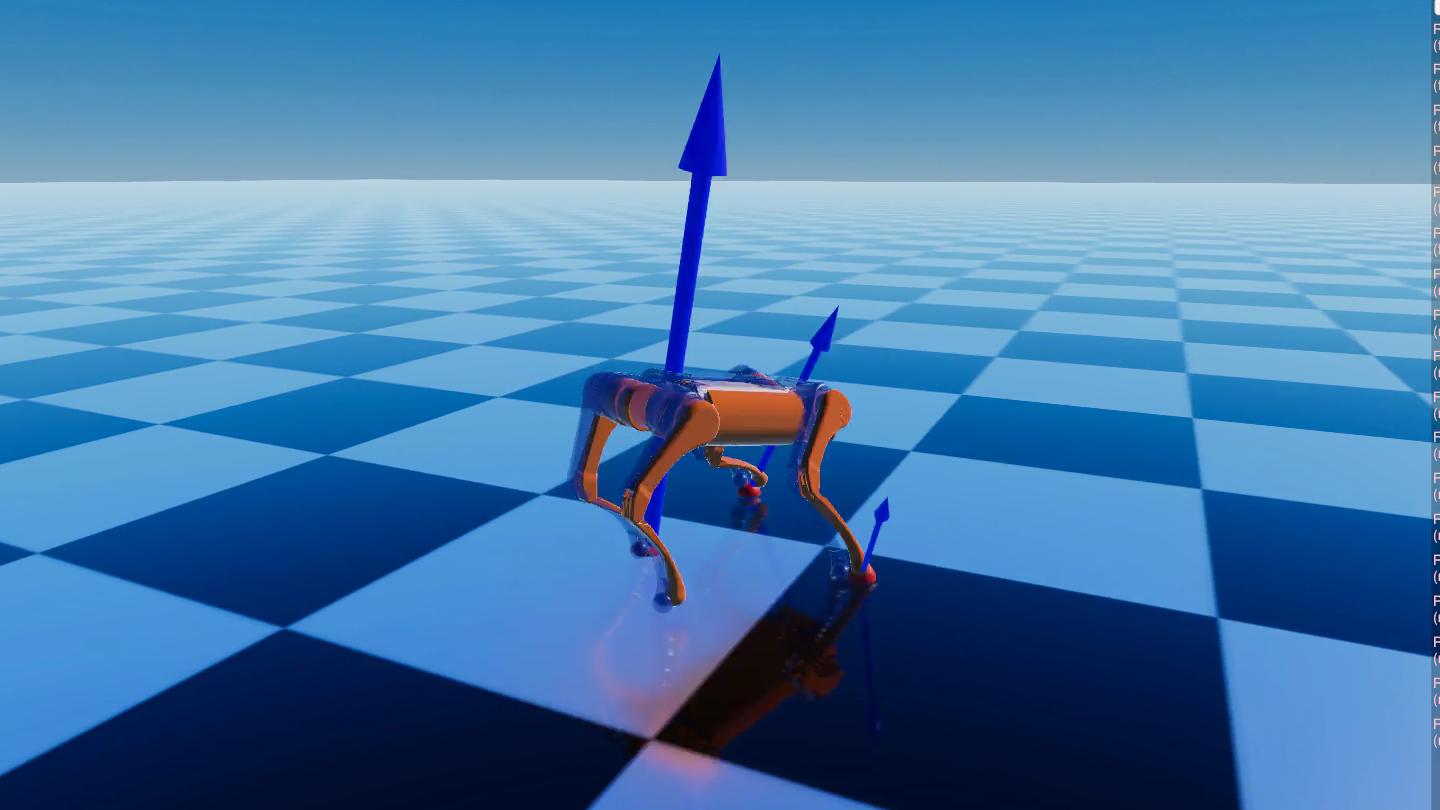} &
    \includegraphics[width=0.195\textwidth]{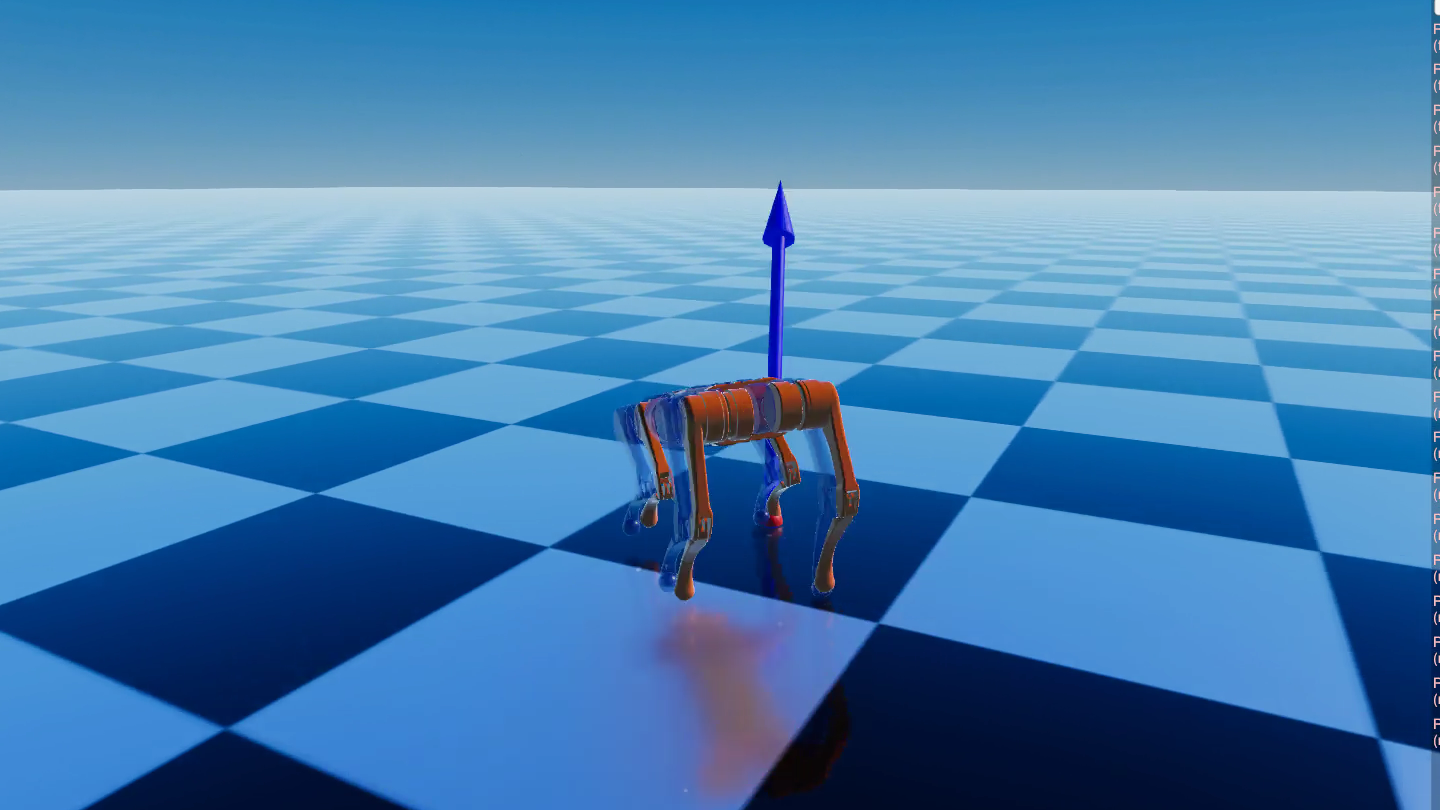} &
    \includegraphics[width=0.195\textwidth]{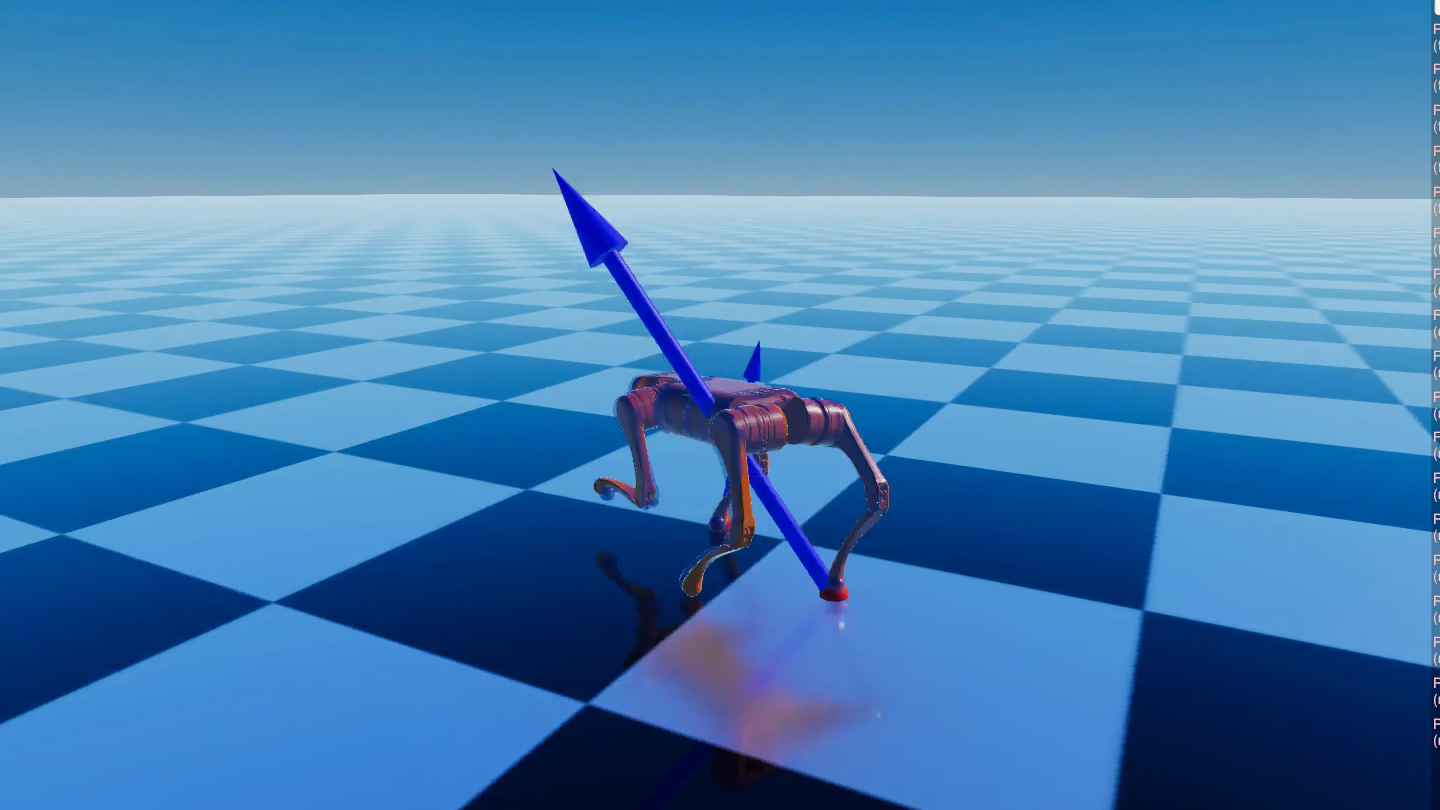} \\
    \includegraphics[width=0.195\textwidth]{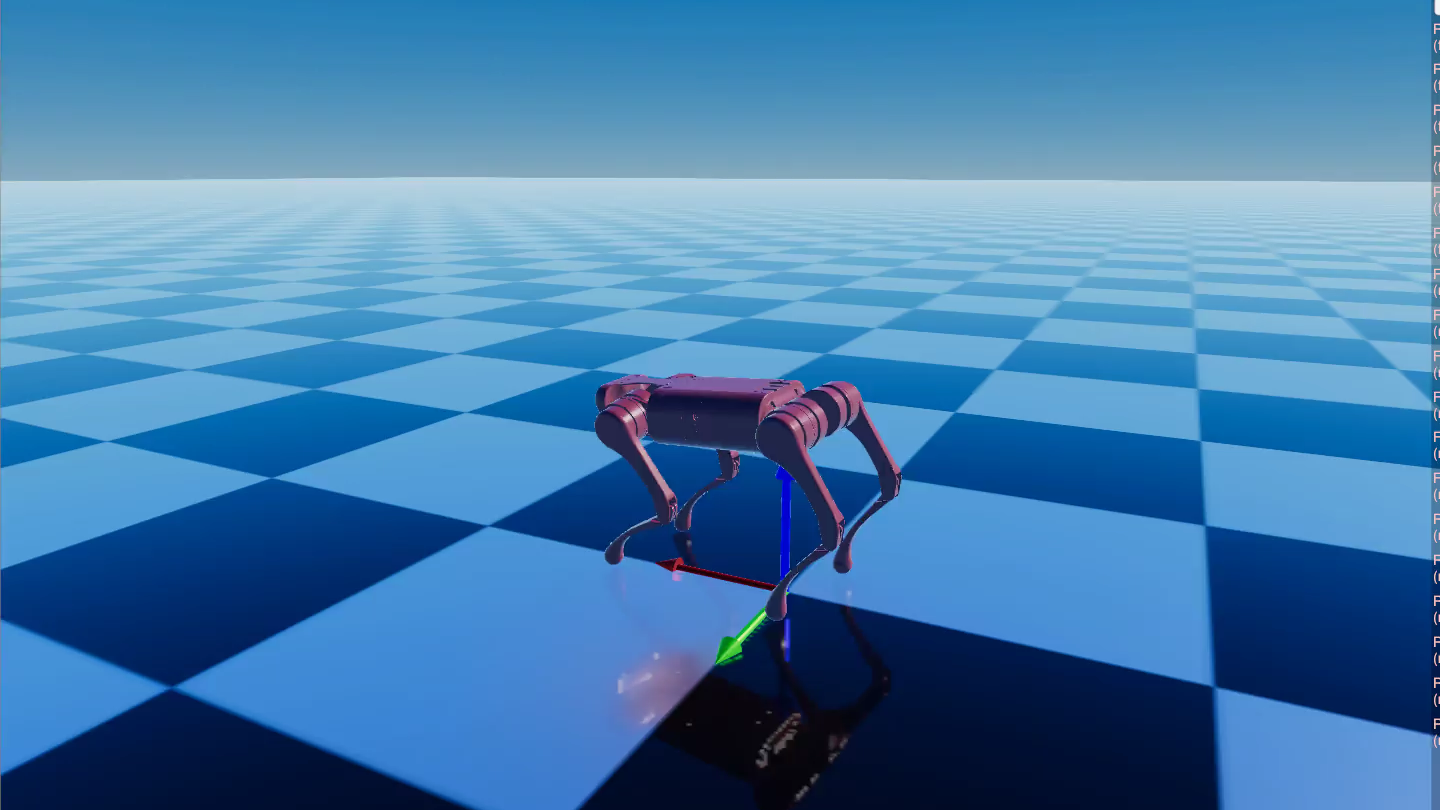} &
    \includegraphics[width=0.195\textwidth]{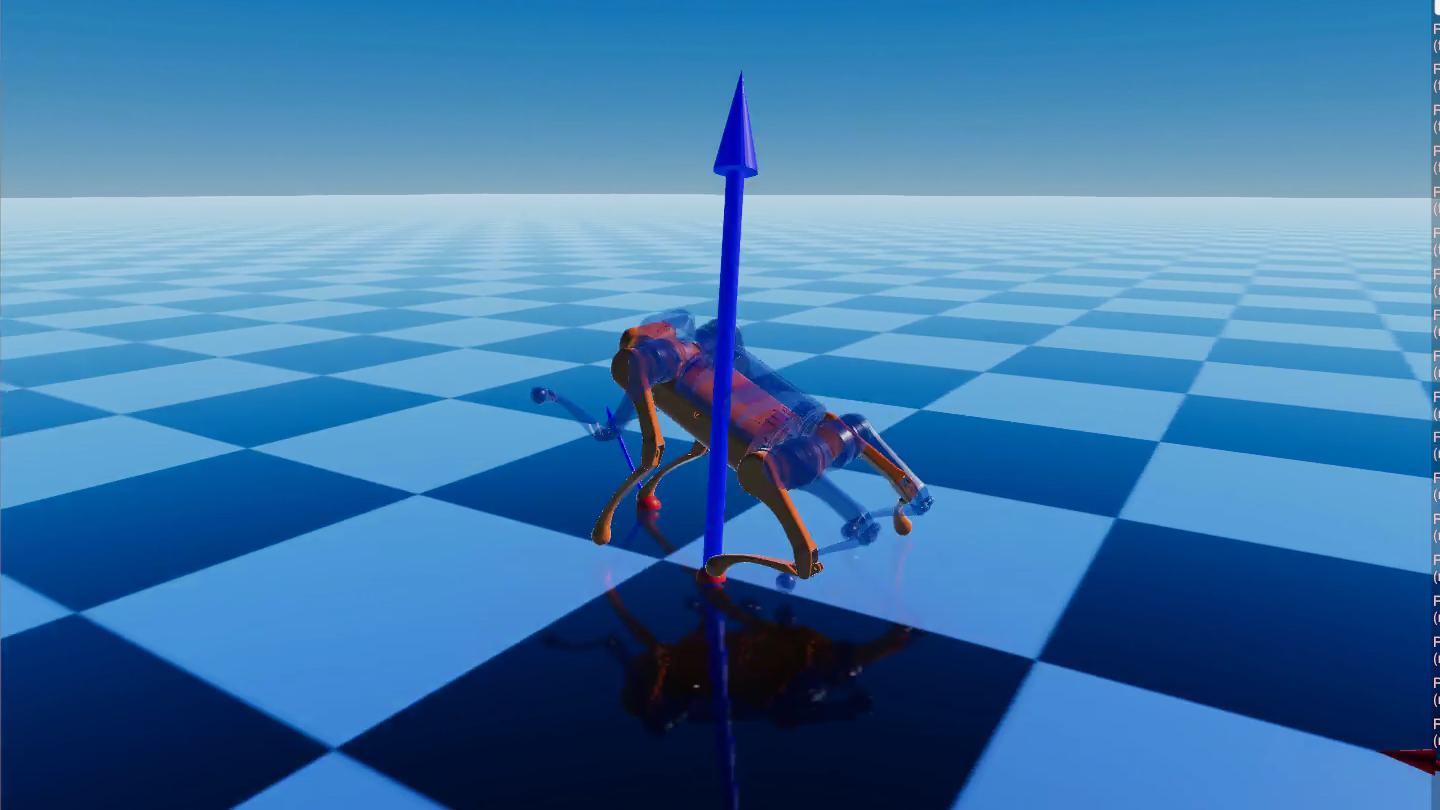} &
    \includegraphics[width=0.195\textwidth]{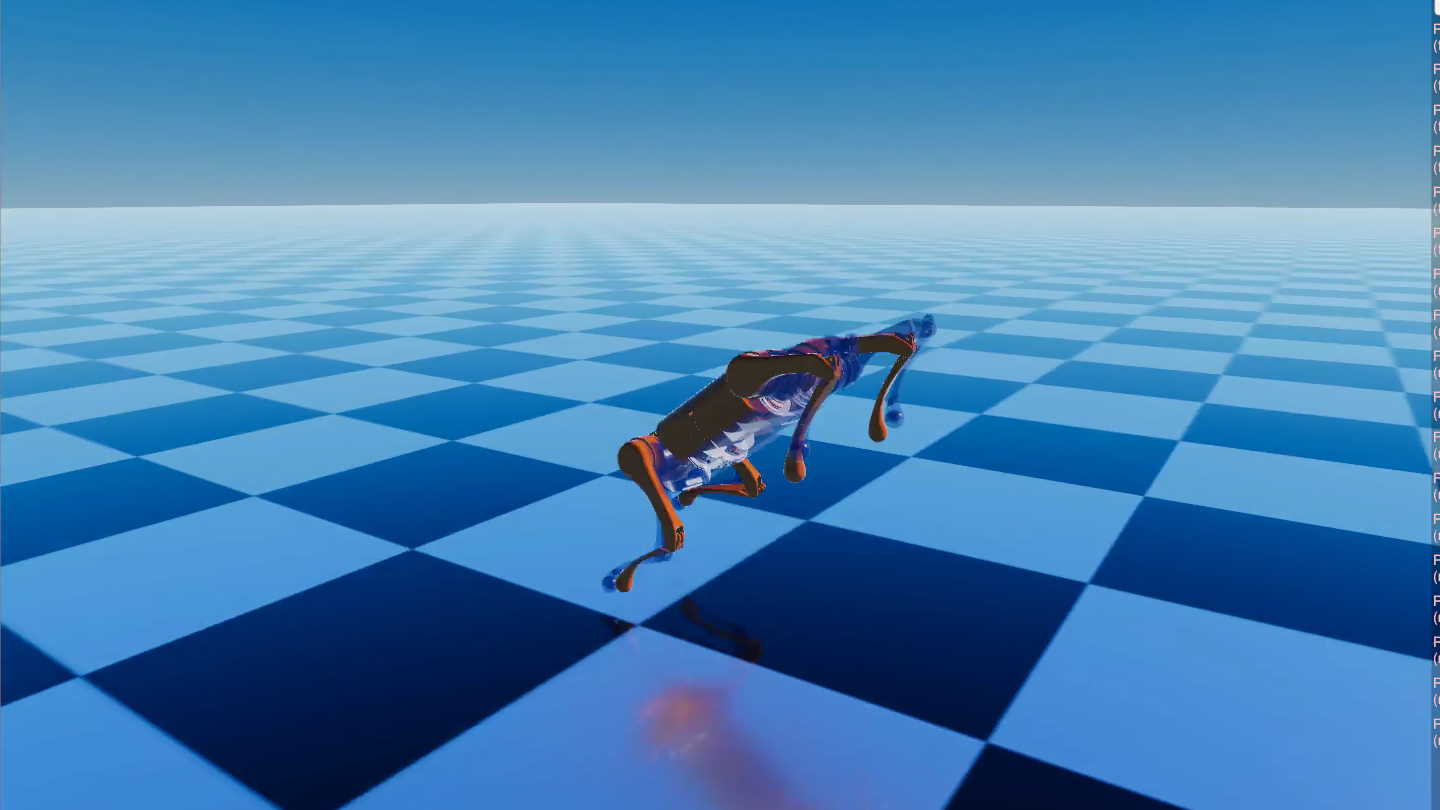} &
    \includegraphics[width=0.195\textwidth]{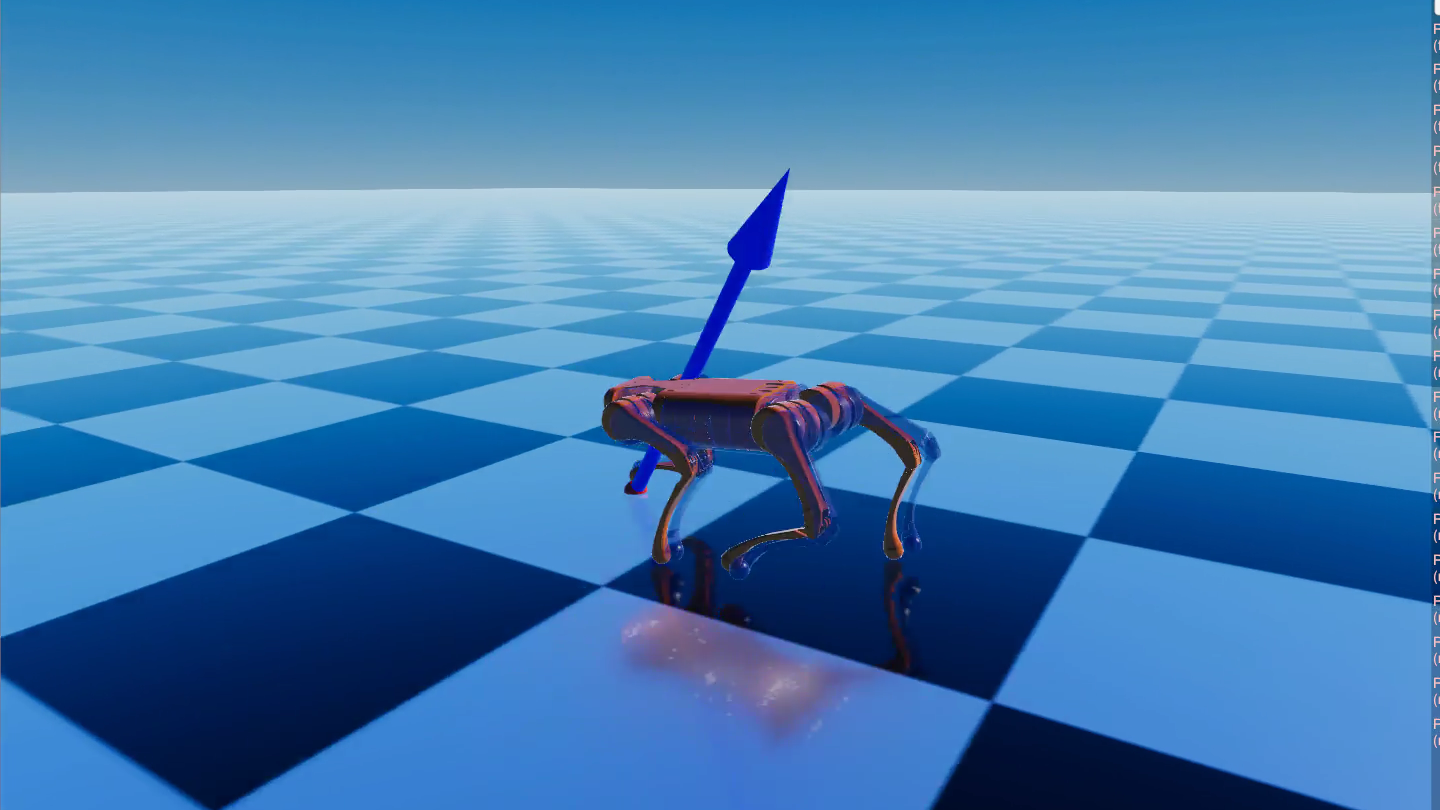} &
    \includegraphics[width=0.195\textwidth]{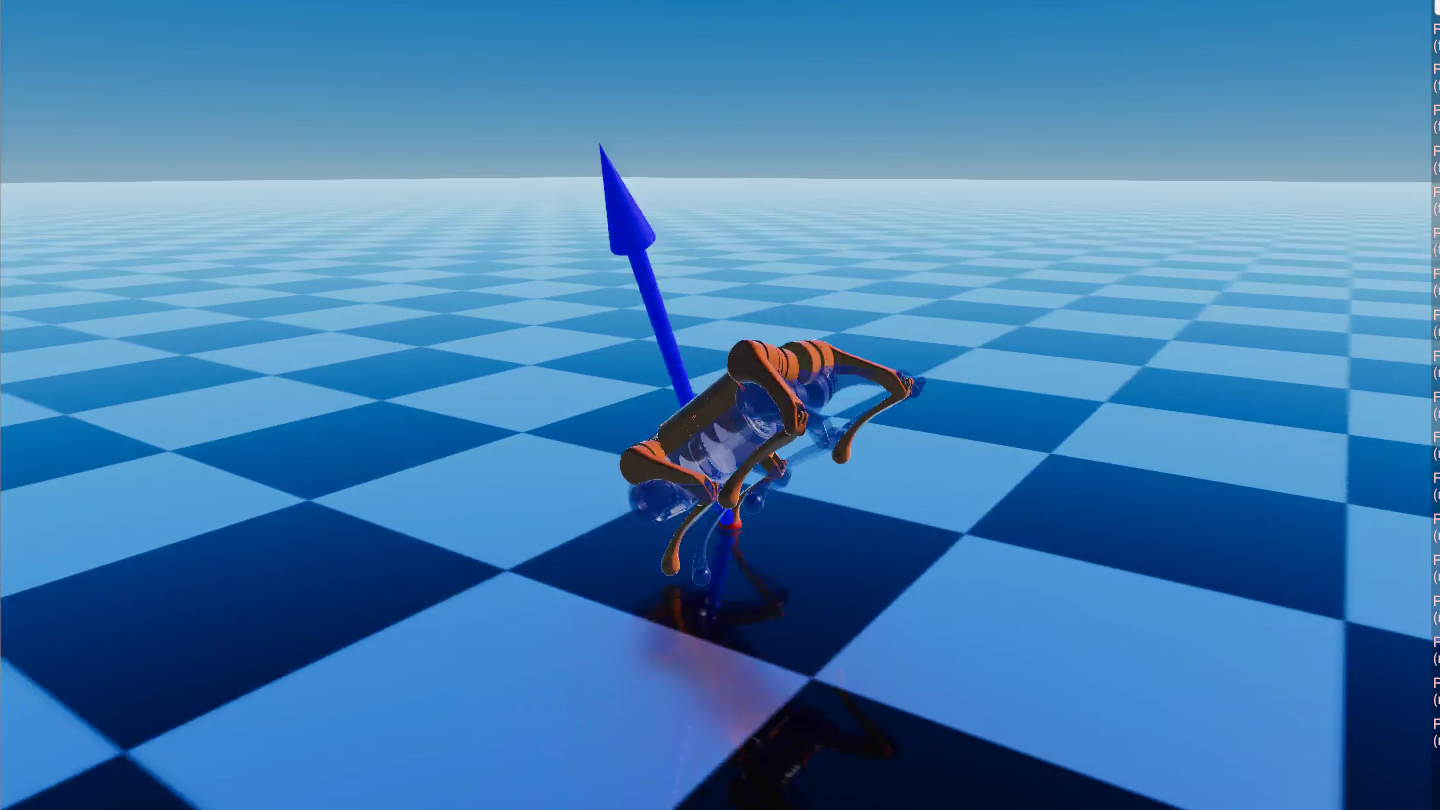} \\
    \includegraphics[width=0.195\textwidth]{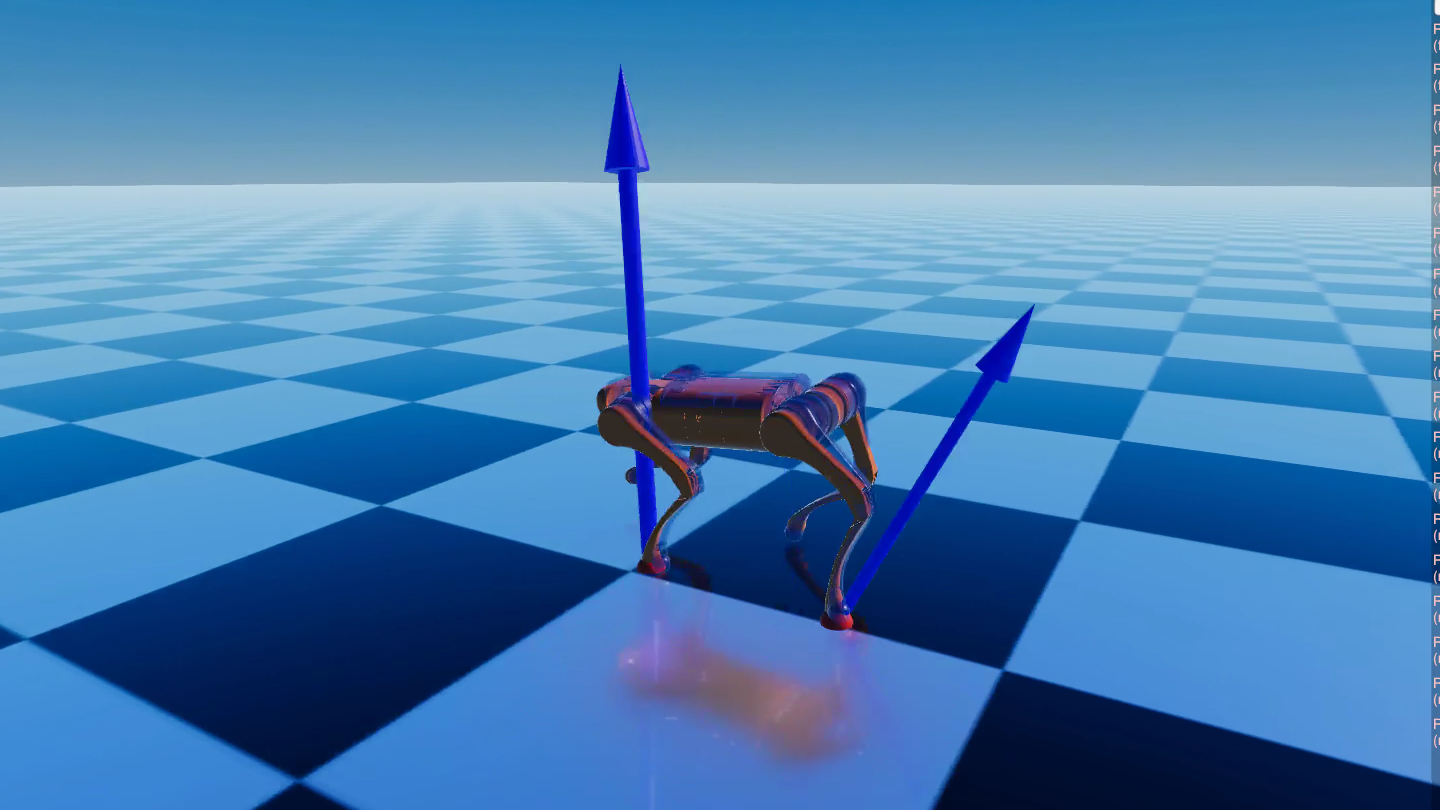} &
    \includegraphics[width=0.195\textwidth]{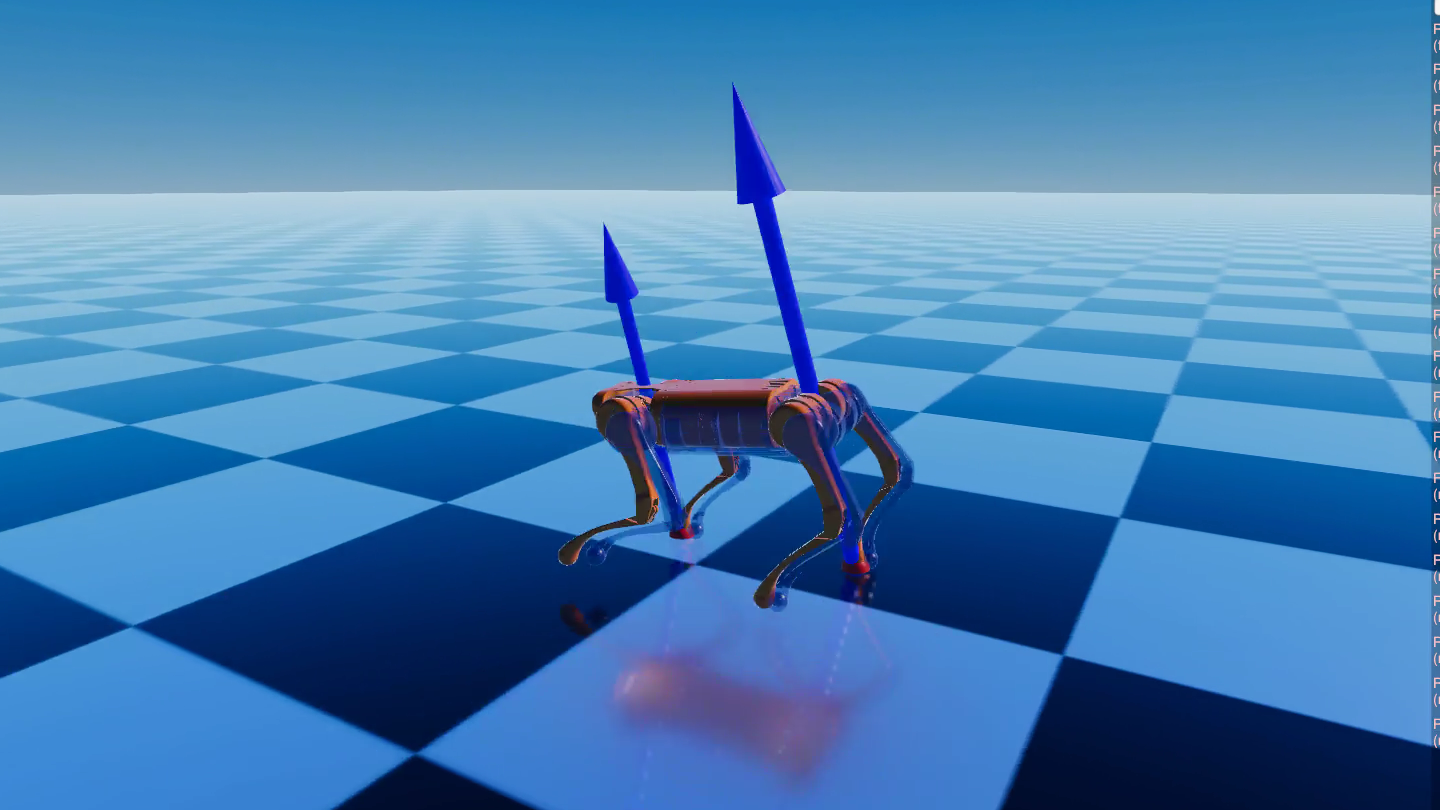} &
    \includegraphics[width=0.195\textwidth]{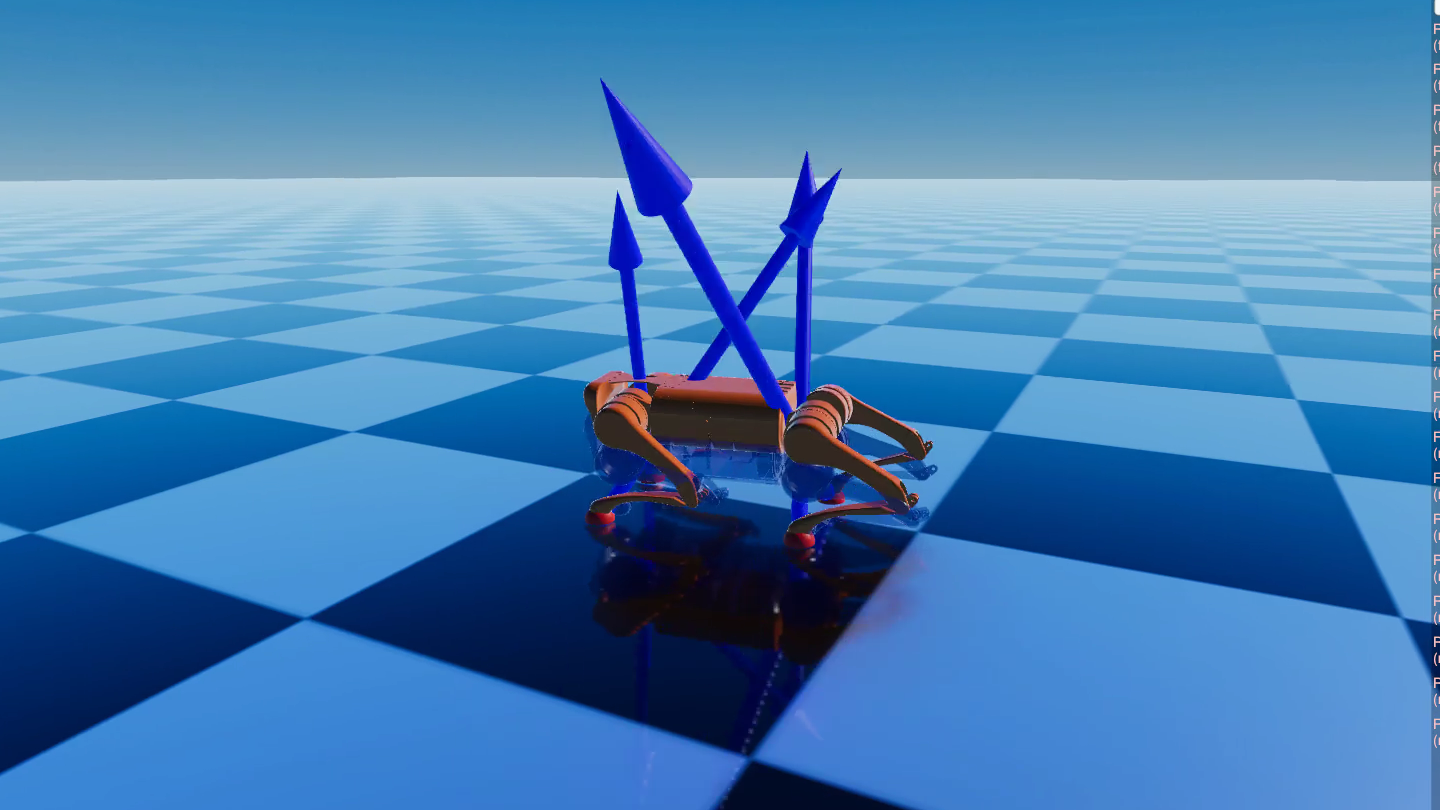} &
    \includegraphics[width=0.195\textwidth]{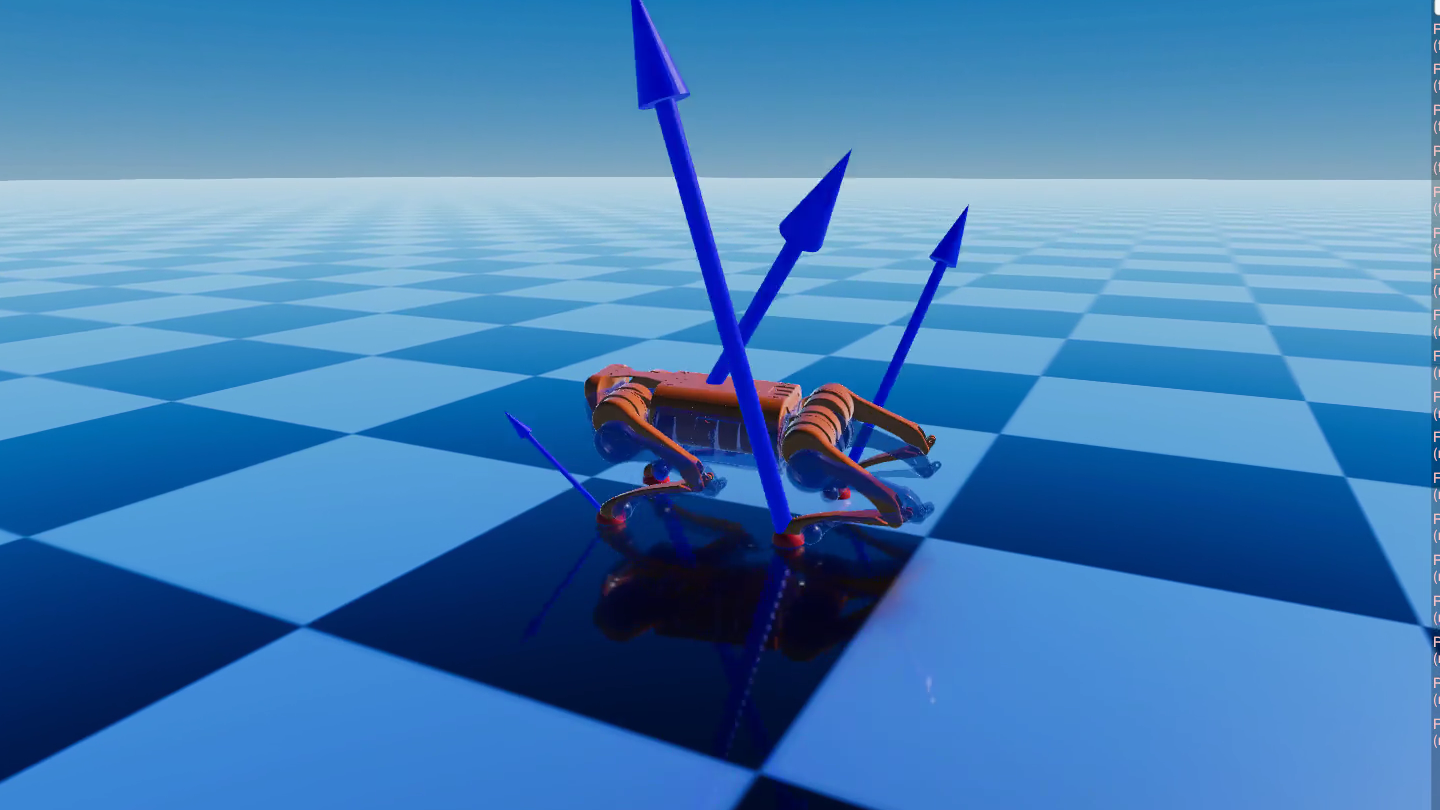} &
    \includegraphics[width=0.195\textwidth]{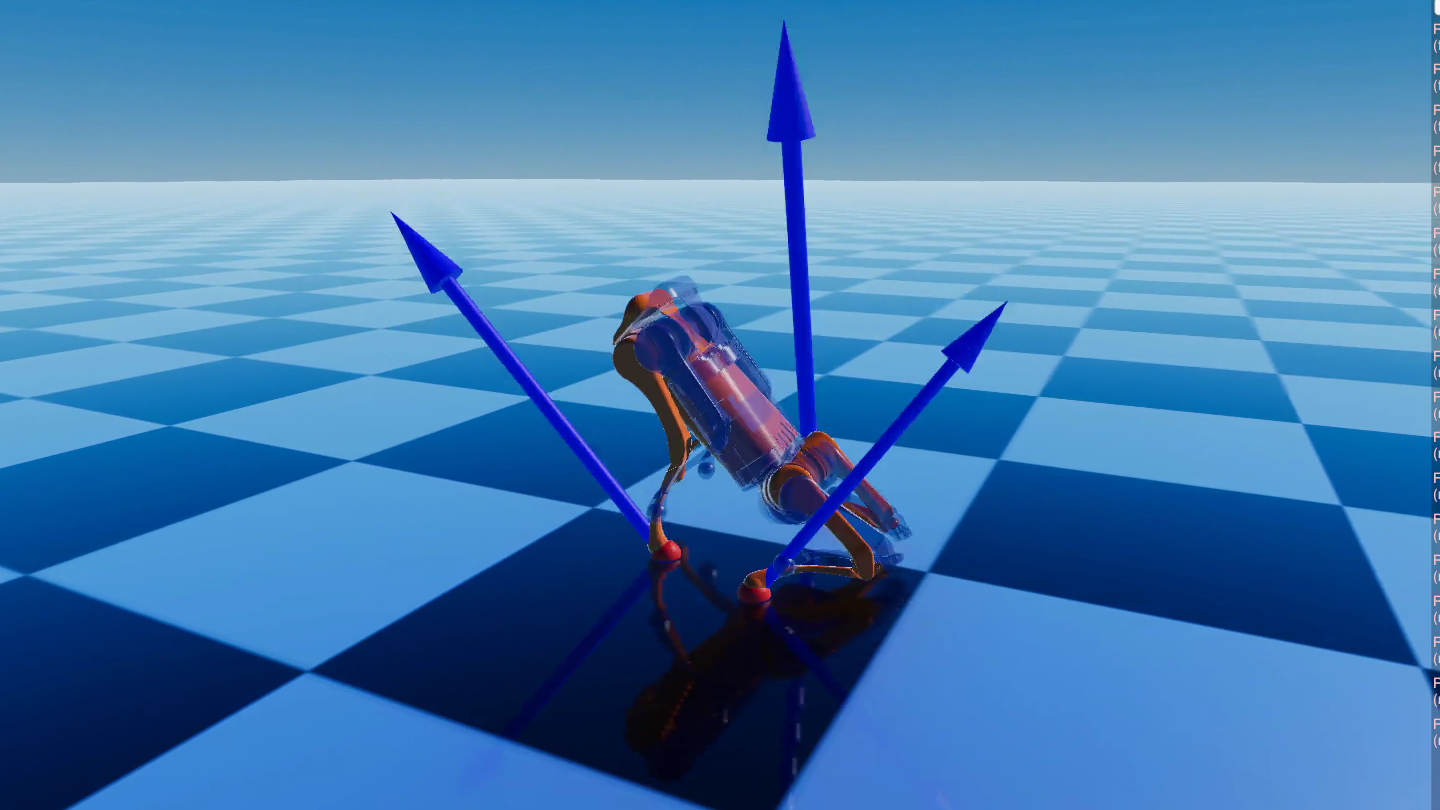} \\
    \end{tabular}
    
    \caption{A variety of generated motions using a single policy. 
    \textbf{1st row:} a sharp turn during a star-trajectory tracking task.
    \textbf{2nd row:} multiple jumps in one sequence.
    \textbf{3rd row:} lying down and sit motions.
    Please refer to the supplemental video for the entire sequences.
    }
    \label{fig:motions}
\end{figure*}

\subsubsection{Early Termination}
In contrast to \cite{deepmimic, won2020}, which uses early termination based on the CoM tracking performance \cite{deepmimic}, or the reward function \cite{won2020} in order to speed up the motion tracking learning, and avoid poor performing tracking policies. Our formulation uses simple contact-based termination without penalty. The only allowable contacts are the four feet with the ground. Won et al.~\cite{won2020} pointed out that contact-based terminations prevented them from learning motions which included self-body contacts. However, as our final goal is to deploy learned behaviors on a real robot, self-body contacts or inadmissible contacts are not desirable. Our formulation, without penalty and prior in the action space (joint residuals w.r.t. a nominal joint configuration) allows the policy from learning certain motions where the kinematic reference has inadmissible contacts, the policy tries to satisfy the high-level reference state (CoM, End-effector, body orientation) instead of trying to reproduce the low-level reference data (joint positions) at any cost.


\subsection{Adaptive Motion Sampling} \label{sec:adaptive_sampling}
Although many researchers have demonstrated successful motion imitation using deep RL, it is still challenging to learn a single policy for various heterogeneous locomotion skills \cite{deepmimic, peng2020learning, Yang_2020}, including walking, turning, and jumping. One difficulty is that the computation of gradients is highly affected by a stochastic sampling of simulation rollouts, which is impossible to cover the entire range if there are too many reference motions in the database. Even worse, the stochastic nature of deep RL can lead a policy to forget about previously learned motor skills, which is referred to as catastrophic forgetting.
\\\indent Our Adaptive Motion Sampling (AMS) allows us to train our policy from an unlabelled and unbalanced dataset. We found that our policy even performed better at tracking the reference motion data and producing natural joint motions when trained directly on all the motion clips at once. No pre-training is thus required. Having a rich set of motions to train on prevents the policy from overfitting on the dynamics and kinematics of certain locomotion skills, serving as data augmentation, and promoting more general locomotion strategies which can track a rich set of motions. As pointed out in \cite{heess2019}, complex skills emerge when trained on a rich set of tasks.
We propose a novel adaptive sampling scheme to overcome these challenges. Our key idea is to maintain two sets of the reference motions: $\mathcal{U}$ and $\mathcal{S}$, which represent unsuccessful and successful motions, respectively. At the beginning of learning, we initially assign all the motions to the unsuccessful group $\mathcal{U}$ and set the successful group $\mathcal{S}=\emptyset$. Sampling from these sets is done following a uniform distribution, and without re-drawing so that after several episodes the policy has been trained on the entirety of the sets. For every $200$ policy iteration, we evaluate the current policy on all the motions and classify them into each group again based on their performance. If the policy is able to track the motion until the end without early termination, we assign the given motion to the successful group $\mathcal{S}$ (resp. $\mathcal{U}$).

Once the motions are classified into two groups, $\mathcal{U}$ and $\mathcal{S}$, we adjust the sampling of the reference motions. We sample $70$~\% of the reference trajectories from $\mathcal{U}$, while taking $30$~\% of trajectories from $\mathcal{S}$. This mechanism allows the policy to majorly focus on difficult reference motions that are yet unsuccessfully learned while not forgetting already learned motions.

\section{Results}

\subsection{Implementation Details}
We develop the proposed framework using RaiSim~\cite{raisim}. We use the integrated implementation of Proximal Policy Optimization~\cite{schulman2017proximal} for learning. Our neural network policy has two layers of [256, 256] hidden neurons with LeakyReLu activation functions. We select an A1 quadrupedal robot~\cite{unitree} from Unitree as an experimental platform. We conduct all the experiments using a desktop with AMD Ryzen Threadripper 3970X 32 cores CPU, and RTX 3090. Using AMS a policy
trained from scratch on 701 motions takes about 40 hours
with 100 environments, 30 threads, and episodes of length
10 seconds. For the motor gains, we choose a proportional
gain $k_p=50.0$ and a derivative gain $k_d=2.0$ in order
to support more stable learning and smoother motions. The
entropy coefficient is chosen as $\epsilon=0.0001$, the policy is
queried every 0.02s and the motion references are played at
a frequency of 1kHz.


\subsection{Generating Diverse Motions}

Our framework is able to learn a single capable policy that can track a large number of trajectories with great diversity, including walking, turning, jumping, sitting, and lying. Using AMS, the policy can successfully track all 701 motions, and $\approx90\%$ of 47 long random mixed motion clips that are used for validation and to test the ability of our policy to generalize to out-of-distribution motions it has never seen. 
\\\indent An episode length is taken as 10 seconds. Most motions last 10 seconds, but for instance, the star motion lasts 40 seconds. Although the policy is only trained for the first 10 seconds of the motion clip, the policy can successfully track the entire star motion, which supports its generalization capabilities. The policy is able to re-use to some extent the learned locomotion skills to track unseen motion references. This makes the learning faster as training on longer motion clips is not required. Instead, it is possible to train on segments or individual locomotion skills present in a longer motion clip.
\\\indent The policy can track motions that contain a lot of transitions between skills, and sudden changes in yaw, or speed for instance. Indeed, it is able to track a short clip that involves lying down and sitting to demonstrate generalization over non-locomotion tasks. Finally, we demonstrate that our policy can track a very long sequence that involves many different components, including walking, turning, different gaits, speed changes, and more. Note that it will be very difficult to develop a single control policy to execute all the motor tasks included in our testing sequences. Please refer to Fig.~\ref{fig:motions} and the supplemental videos for qualitative evaluation. We will also provide more quantitative analysis in the following section.



\section{Analysis} \label{sec:analysis}


\subsection{Influence of Past and Future Target Information in Observations}
\begin{figure}
\centerline{\includegraphics[width=0.47\textwidth]{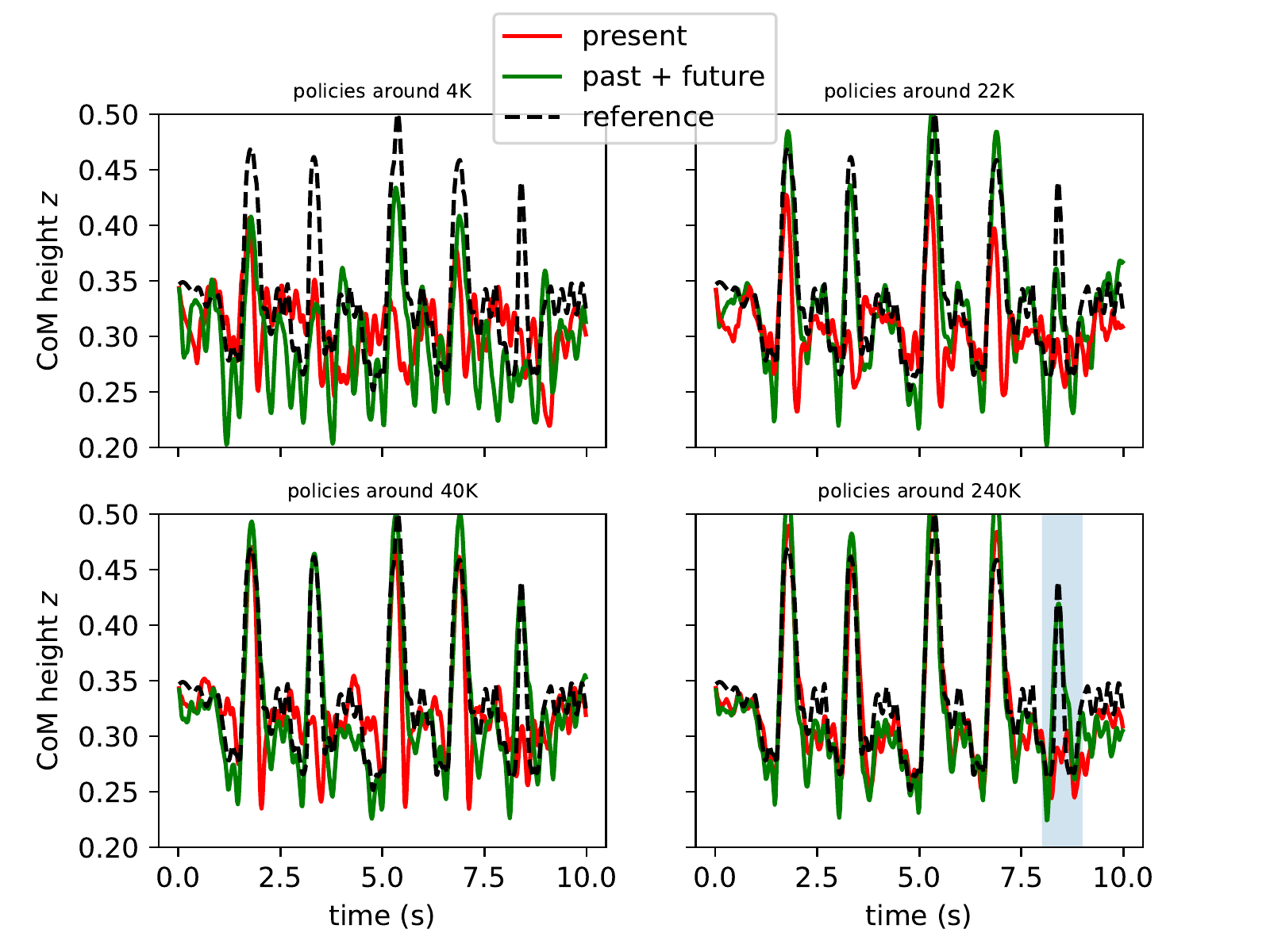}}
\caption{Influence of the past and future trajectory. Our observation space that includes the past and future enables faster learning: even at the 22K-th policy iteration (top right), our design (green) tracks the reference motion better than the baseline design (red). It also results in a better final policy. At the 240K-th policy iteration (bottom right), our design can track all five jumps while the baseline misses the last jump as highlighted in the blue box centered at around 8.0 seconds.}
\label{fig-ref-in-obs}
\end{figure}
Inspired by other works~\cite{won2020, peng2020learning}, the policy is provided with future and past information of the reference to track. The current reference frame is excluded in order to incentivize the agent to interpolate between the different keyframes in order to prevent the policy from overfitting on the low-level kinematic reference data. Fig. \ref{fig-ref-in-obs} shows that this configuration encourages a faster learning and higher quality learned (smoother, more symmetric) joint trajectories. Fig. \ref{fig-ref-in-obs} presents the CoM height tracking for a policy with only the present reference information and a policy with past and future information. First, we find that our design of past and future reference converges faster, as green curves (ours) are closer to the blue-dotted reference motion than red curves (the baseline design with only the current frame) in early policy iterations. Even after a long-enough training with 240k iterations, the policy with the baseline observation design misses the final jump and decides to run through. We hypothesize that past and future trajectory information is critical to plan ahead dynamic jumping motions.

 \begin{figure}
\centerline{\includegraphics[scale=0.36]{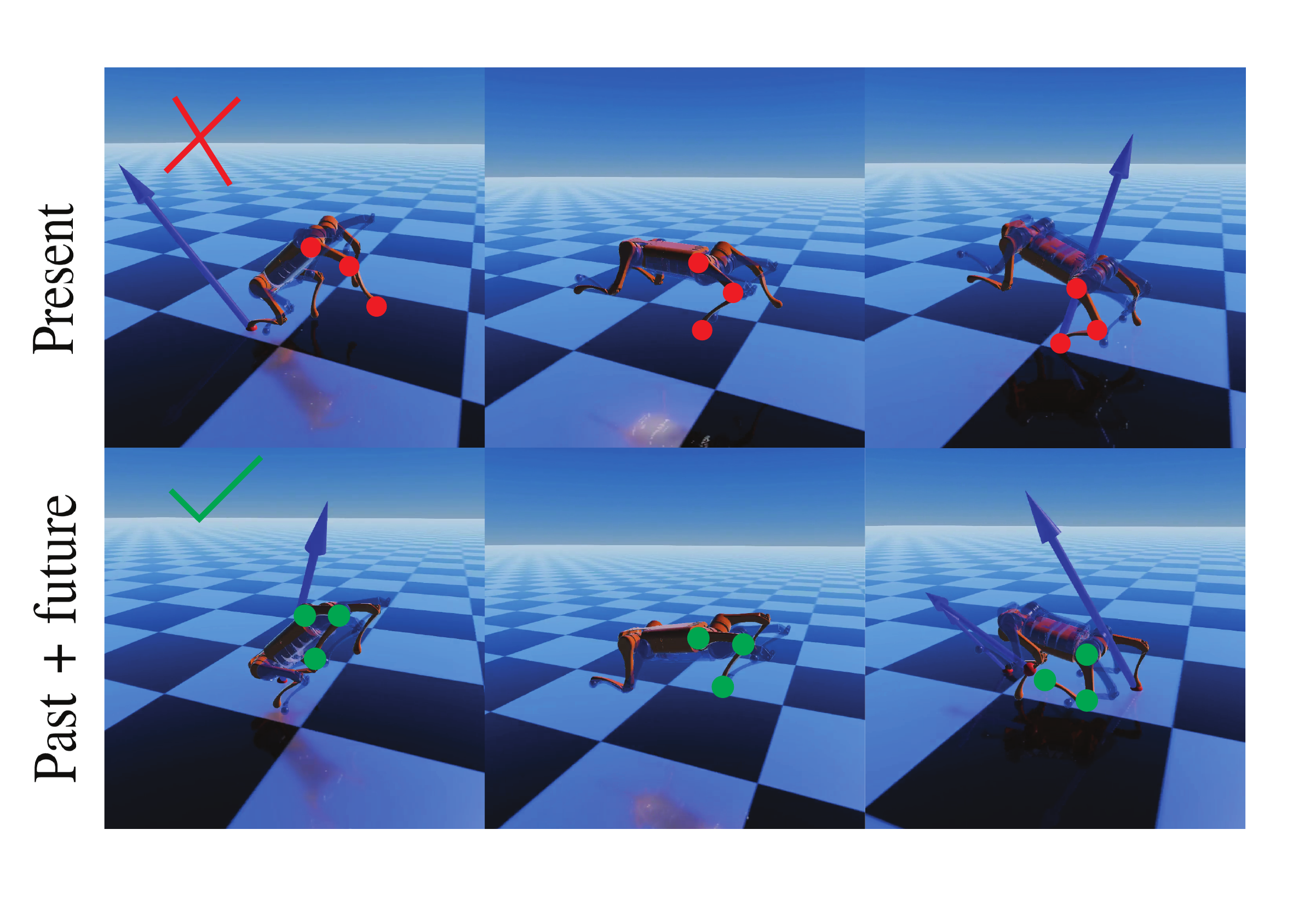}}
\caption{Selected key frames of a dynamic jump tracking: the fastest jump present in the dataset at a maximum CoM speed of $1.76 m/s$. 
The baseline policy shows reactive and awkward behaviors for balancing (\textbf{top}) while our observation design leads to more natural and smooth jumping motions (\textbf{bottom}) .}
\label{fig-jump-key-frames}
\end{figure}

A qualitative comparison is presented in Fig. \ref{fig-jump-key-frames}. The baseline observation with only the current frame leads to a reactive behavior that uses its rear legs awkwardly in order to balance. In fact, we observe this behavior consistently over all 200 jumping motions the baseline was trained on. On the other hand, our observation space with past and future reference information finds smooth and natural jumping behaviors which can handle even multiple jumps, and even exhibits alterations from the kinematic reference that look as natural as the reference (See Fig. \ref{fig-jump-key-frames}).

\subsection{Influence of Action Prior}

\begin{figure}
\centerline{\includegraphics[width=0.4\textwidth]{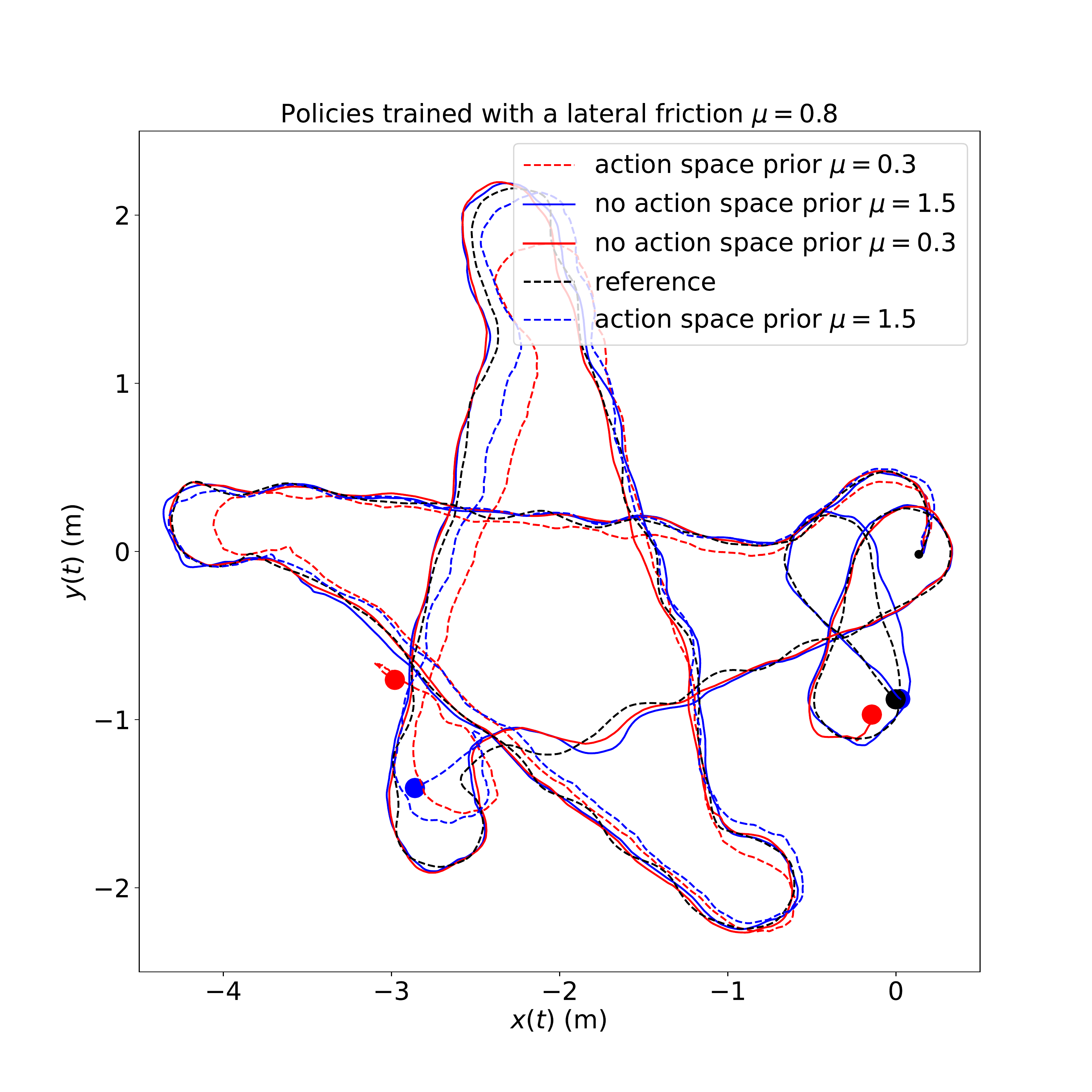}}
\caption{Generalization over unseen frictions of 0.3 (red) and 1.5 (blue). We examine policies with and without action prior. The policies without action prior (ours, solid lines) show better robustness, while the policy with action prior (baseline, dotted lines) shows larger tracking errors and even cannot complete the sequence. Circles represent the location where the episodes end.}
\label{fig-robust-friction}
\end{figure}

There exist two common choices of action spaces in the literature of motion imitation. The first is to define it as the delta to the current frame of the reference motion under the expectation that the desirable PD targets are closer to the reference motion (\emph{Action Prior}). The second is to make it independent from the reference motion, such as the delta to the fixed nominal pose (\emph{No Action Prior}, ours). In our experience, a policy without action prior shows much better robustness, particularly when it is combined with our joint-agnostic reward design (Eq.~(\ref{eq:reward})).

Fig.~\ref{fig-robust-friction} illustrates well the generalization capability of policies over unseen surfaces with low (0.3, red) and high (1.5, blue) lateral friction coefficients ($\mu$), whereas the policy is trained with $\mu=0.8$. In our experience, learning with \emph{action priors} overfits to track the joint motions and does not generalize well when the robot starts to deviate from the desired trajectory. As a result, the policy exhibits high tracking errors (dotted lines) and even terminates early. On the other hand, our learning formulation without both action prior and joint tracking objective allows the policy to show more robust behaviors to complete complex star-shaped trajectories (solid lines).



\subsection{Adaptive sampling}

Finally, this subsection analyzes the importance of our adaptive motion sampling (AMS). In our experience, AMS is critical to cover a large number ($\sim700$) motions without missing a few outlier motions, such as jumping, lying, and sitting. For instance, we have $200$ pace motions while having only $10$ of jumping motions. Therefore, naive sampling will likely prioritize pace motions. 

\begin{figure}[htbp]
\centerline{\includegraphics[width=0.47\textwidth, height=0.3\textwidth]{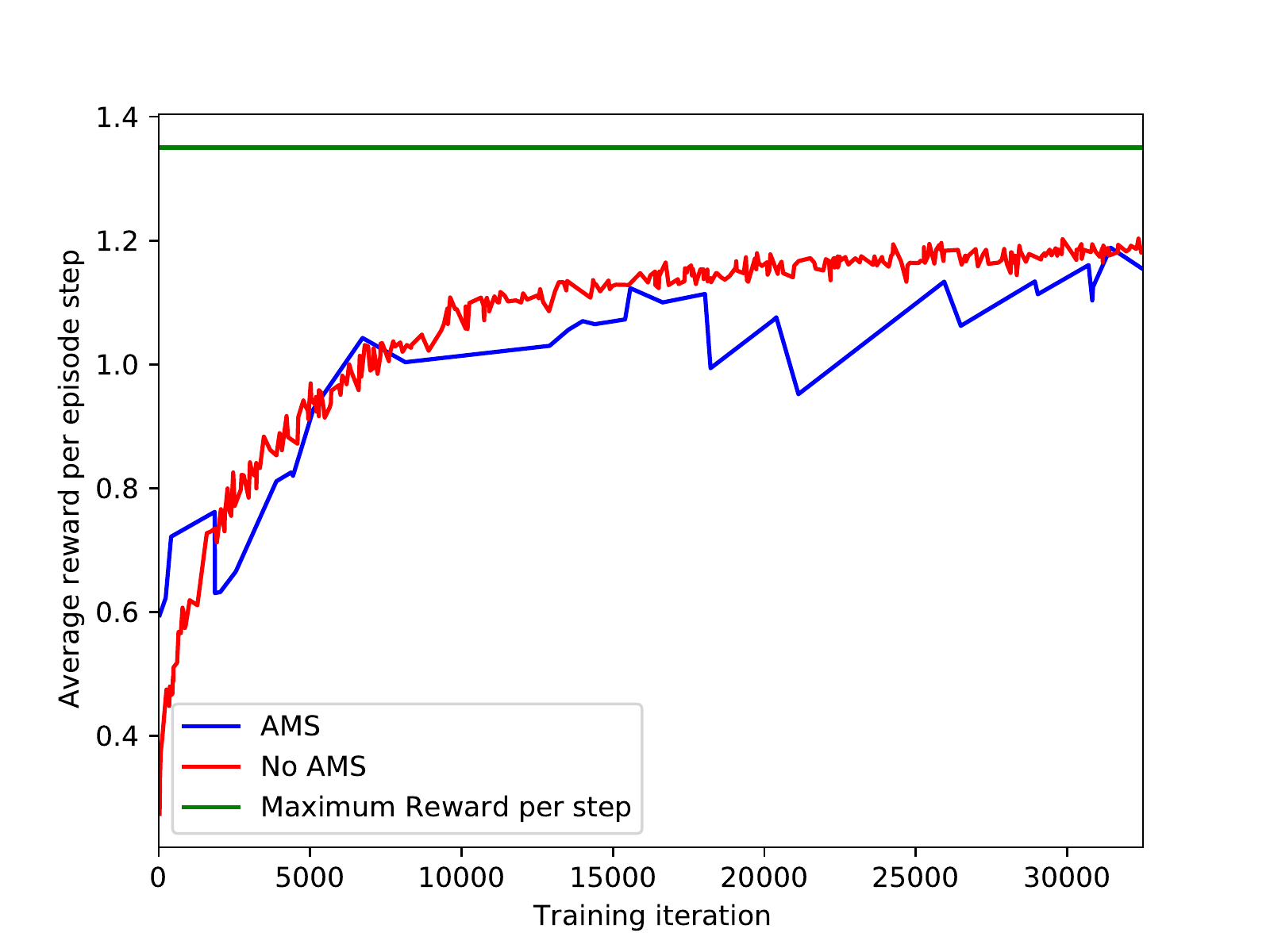}}
\centerline{\includegraphics[width=0.47\textwidth,, height=0.3\textwidth]{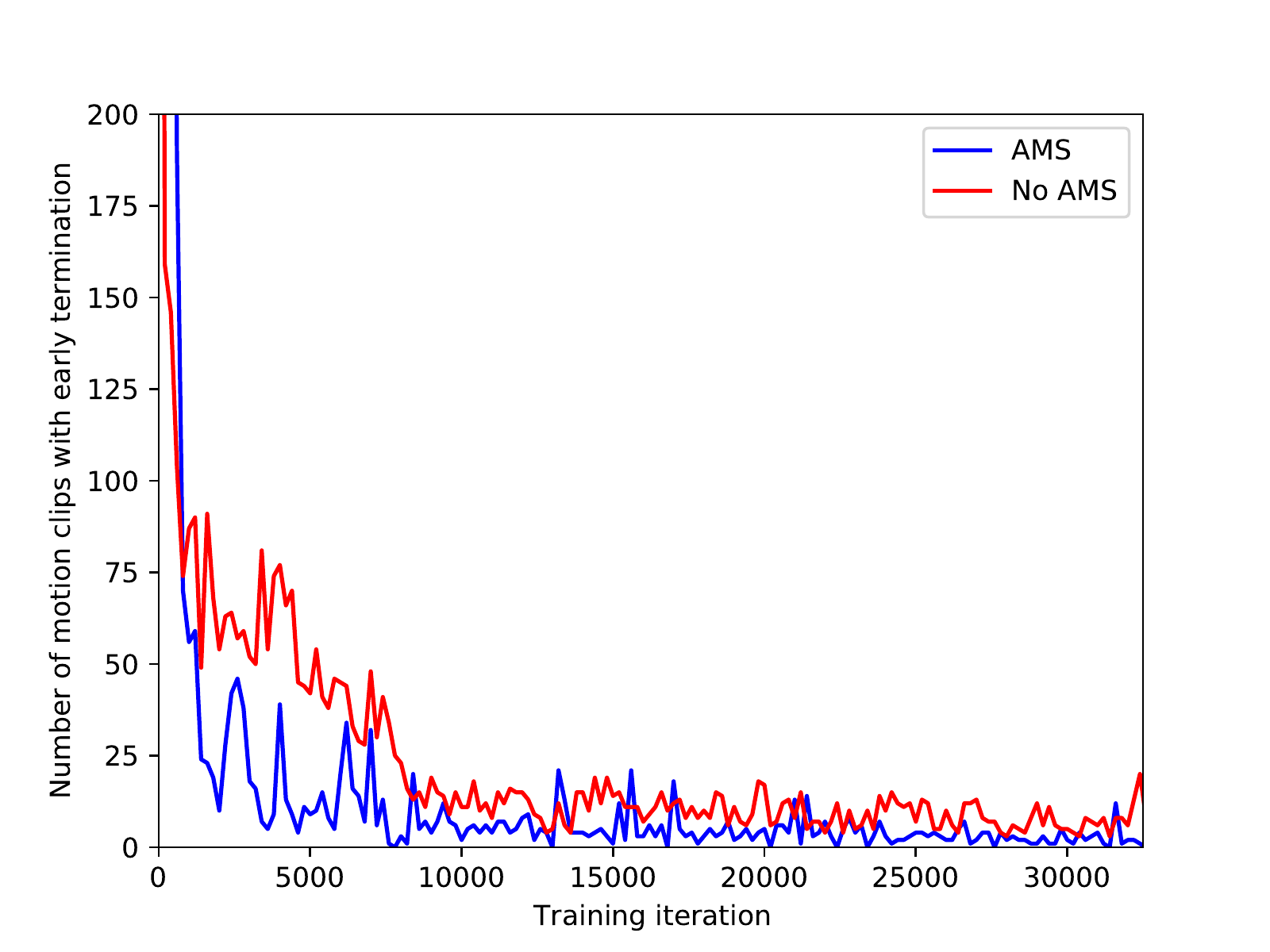}}
\caption{Adaptive Motion Sampling (AMS): comparison of training results with a policy trained with AMS and one policy without AMS. AMS looks suboptimal in terms of the episodic reward (\textbf{top}), but it actually successfully tracks a lot more motions than the baseline without AMS (\textbf{bottom}).}
\label{fig-ads-training-curve}
\end{figure}

We plot (1) the average episodic reward over time and (2) the number of failed motions in Fig.~\ref{fig-ads-training-curve}. From the perspective of the conventional reward curve (top), it seems that AMS performs suboptimally with slightly lower episodic rewards. However, please note that AMS puts a policy in tougher scenarios by sampling harder tracking problems more often, and we cannot directly compare the reward function. Therefore, we also plot the number of the failed trajectories out of $701$ motions at the bottom of Fig.~\ref{fig-ads-training-curve}, as a more fair comparison criterion. It shows that our AMS fails less over by not ignoring some minority motions.

\section{Conclusion}
We present a scalable motion imitation framework to learn a single policy that can track a large variety of motions, including walking, turning, running, jumping, sitting, and lying. Starting from the existing motion imitation framework~\cite{deepmimic} , we carefully design the observation space, action space, and reward function to improve the effectiveness and robustness of the final policy. In addition, we propose an adaptive motion sampling scheme, which is designed to focus on the learning of more challenging trajectories and to avoid catastrophic forgetting of the previously learned skills. We successfully train a very versatile single policy from a large number of trajectories. We demonstrate that it can also  generalize well to novel trajectories to execute a complex, long motion sequence that involves many different motor skills. In addition, we also showcase that the learned policy is robust against the change of environment parameters such as lateral friction. We finally analyze the importance of our problem formulation and adaptive motion sampling by conducting a set of experiments.

There exist several interesting future research directions that we want to explore. For instance, we plan to add more reference motions to the database for non-trivial tasks, such as stair climbing, crawling, and walking over rough terrains. However, the current data generation scheme of motion retargeting will have limitations because it relies on the existing public motion capture data set of a real dog. One possible solution is to add more data using the off-the-shelf trajectory optimization framework~\cite{winkler2018gait}, which can generate physically valid trajectories for various environments. Once we increase the size of the database, we may need to even further improve the scalability of the current learning framework. It can be approached by adopting more parallelized reinforcement learning algorithms~\cite{wijmans2019dd}, investigating novel policy architecture~\cite{kumar2022cascaded}, structuring the dataset~\cite{won2020} or adopting the framework of adversarial learning~\cite{peng2021amp}.

Our obvious next step is to deploy the learned policy on the real hardware of the A1 robot. We expect that the learned policy needs to cross a large sim-to-real gap, which can be approached by system identification or domain randomization. However, domain randomization may also increase the difficulty of the problem, and the learning may not converge effectively. In this case, we may want to pretrain a policy without domain randomization and fine-tune the policy in randomized environments.

\bibliographystyle{ieeetr}
\bibliography{ref}

\end{document}